\documentclass[12pt]{spieman}  
\usepackage{amsmath,amsfonts,amssymb}
\usepackage[pdftex]{graphicx}
\usepackage{lipsum}
\usepackage{lscape}
\DeclareGraphicsExtensions{.pdf,.jpeg,.png}
\usepackage[ruled]{algorithm2e}
\usepackage{rotating}
\usepackage{url}
\usepackage{graphicx}
\usepackage{afterpage}
\usepackage{setspace}
\usepackage{tocloft}
\usepackage{epstopdf}
\usepackage[dvipsnames]{xcolor}
\usepackage{bm}
\usepackage{lineno}

\title{Reproducibility of an airway tapering measurement in CT with application to bronchiectasis.}

\author[a,*]{Kin Quan}
\author[a]{Ryutaro Tanno}
\author[b]{Rebecca J. Shipley}
\author[c]{Jeremy S. Brown}
\author[a,c]{Joseph Jacob}
\author[c]{John R. Hurst}
\author[a]{David J. Hawkes}
\affil[a]{Centre for Medical Image Computing, University College London, Gower Street, London, UK, WC1E 6BT}
\affil[b]{Department of Mechanical Engineering, University College London, Gower Street, London, UK, WC1E 6BT}
\affil[c]{UCL Respiratory, University College London, Gower Street, London, UK, WC1E 6BT}

\renewcommand{\cftdotsep}{\cftnodots}
\newcommand{\argmax}{\operatorname{Argmax}}

\newcommand{\deletetext}[1]{}
\newcommand{\newtext}[1]{#1}
\newcommand{\persontext}[1]{}
\newcommand{\replacetext}[2]{\newtext{#2}}
\newcommand{\secondnewtext}[1]{\newtext{#1}}

\begin{document} 
\maketitle

\begin{abstract}\\
	\textbf{Purpose:} This paper proposes a pipeline to acquire a scalar tapering measurement from the carina to the most distal point of an individual airway visible on CT. We show the applicability of using tapering measurements on clinically acquired data by quantifying the reproducibility of the tapering measure. \textbf{Methods:} We generate a spline from the centreline of an airway to measure the area and arclength at contiguous intervals. The tapering measurement is the gradient of the linear regression between area in log space and arclength. The reproducibility of the measure was assessed by analysing different radiation doses, voxel sizes and reconstruction \replacetext{algorithms}{kernel} on single timepoint and longitudinal CT scans and by evaluating the effct of airway bifurcations. \textbf{Results:} Using 74 airways from 10 CT scans, we show a statistical difference, p = $3.4 \times 10^{-4}$ in tapering between healthy airways (n = 35) and those affected by bronchiectasis (n = 39). The difference between the mean of the two populations was 0.011mm$^{-1}$ and the difference between the medians of the two populations was 0.006mm$^{-1}$. The tapering measurement retained a 95\% confidence interval of $\pm0.005$mm$^{-1}$ in a simulated 25 mAs scan and retained a 95\% confidence of $\pm0.005$mm$^{-1}$ on simulated CTs up to 1.5 times the original voxel size. \textbf{Conclusion:} \secondnewtext{We have established an estimate of the precision of the tapering measurement and estimated the effect on precision of simulated voxel size and CT scan dose. We recommend that the scanner calibration be undertaken with the phantoms as described, on the specific CT scanner, radiation dose and reconstruction algorithm that is to be used in any quantitative studies.} \\
	
	\noindent Our code is available at \url{https://github.com/quan14/AirwayTaperingInCT}\\
	
	\noindent The manuscript was originally published as:
	Kin Quan, Ryutaro Tanno, Rebecca J. Shipley, Jeremy S. Brown, Joseph Jacob, John R. Hurst, David J. Hawkes, "Reproducibility of an airway tapering measurement in computed tomography with application to bronchiectasis," J. Med. Imag. 6(3), 034003 (2019), \url{https://doi.org/10.1117/1.JMI.6.3.034003}. 
\end{abstract}


{\noindent \footnotesize\textbf{*}Corresponding author: Kin Quan,  \linkable{kin.quan.10@ucl.ac.uk} }

\begin{spacing}{2}   

\section{Introduction and Purpose}

Bronchiectasis is defined as the permanent dilatation of the airways. Patients with bronchiectasis can suffer severe exacerbations requiring hospital admission and have a poorer quality of life\cite{Chalmers2018}. Clinicians diagnose bronchiectasis on computed tomography (CT) imaging by visually estimating the diameter of the airway/bronchus and its adjacent pulmonary artery and calculating the broncho-arterial (BA) ratio. A BA ratio greater than 1 indicates the presence of bronchiectasis\cite{Pasteur2010}.

Various groups have proposed methods to automatically and semi automatically compute the BA ratio for bronchiectatic airways\cite{Mumcuoglu2013,Perez-Rovira2016,Fetita2015}. However, use of the BA ratio to diagnose bronchiectasis has two major flaws. First of all, the maximum healthy range of the BA ratio can be 1.5 times size of the artery\cite{Hansell2010}. Second, blood vessels can change size as a result of factors including altitude\cite{Kim1997}, patient age\cite{Matsuoka2003} and smoking status\cite{Diaz2017}. This conflicts with the assumption that the pulmonary artery is always at a constant size.

An alternative approach to diagnose and monitor bronchiectatic airways is to analyse the taper of the airways i.e. the rate of change in the cross-sectional area along the airway\cite{Pasteur2010}. In patients with bronchiectasis, the airway is dilated and so the tapering rate must be reduced. Airway tapering is difficult to assess visually and to measure manually. As described by Hansell\cite{Hansell2010}, the observer would have to make multiple cross-sectional area measurements along the airway. \replacetext{Furthermore, this is only possible in airways that run near perpendicular to the CT image plane.}{As mentioned in Cheplygina et al\cite{Cheplygina2016}, measuring multiple lumen is a manually exhaustive task and prone to mistakes.}

\subsection{Related Work}

There have been various strategies to quantify tapering in the airways. The initial proposed tapering measurements by Odry et al.\cite{Odry2006} were restricted to short lengths of the airways. A segmented airway would be spilt into four equal parts. Each segment had an array of computed lumen diameters. The tapering was measured as the linear regression of the lumen diameters along the branch. The method shared similarity to Venkatraman et al.\cite{Venkatraman2006}, but the diameter measurements were taken across the central half of each branch. Various analyses attempted to measure the taper of airways containing multiple branches. In Oguma et al.\cite{Oguma2015} the region of interest from the carina to the fifth generation airway was measured, however this was only performed in patients with COPD. Finally, Weinheimer et al.\cite{Weinheimer2017} used a graphical model of the airways for their proposed tapering measurement. The graphical model was based on a graphical tree originating at the trachea and extending into distal branches, depending on airway bifurcations. A tapering measure was assigned to the edge of the graph depending on the lumen area and generation. They also proposed a regional tapering measurement based on the segments of the lobes of the lungs.

The described tapering measurements have two key limitations. First of all, there is no detailed quantification of reproducibility when considering differences in specifications of the CT scanner, or reconstruction \replacetext{algorithm}{kernel}, making it difficult to compare tapering statistics from  different machines or from the same CT scanner employing different scanning parameters. Second, the region of interest for the tapering measurement was restricted to airways that were segmented using the respective airway segmentation software. Bronchiectasis is a heterogeneous disease –- it can affect any area in the lung including the peripheral regions\cite{Chalmers2015a}. Thus, to encapsulate the disease in the tapering measurement, one would need to consider the region of interest as the entire airway, from the trachea to the most distal point.

In all the proposed tapering measurements, obtaining the cross-sectional area is a necessary input for the algorithms. There have been various analyses attempting to validate the reproducibility and precision of measurements against dose\cite{Hammond2017, Leutz-Schmidt2017,Jia2018}, voxel size\cite{Achenbach2009} \persontext{Changed citation}, reconstruction kernel\cite{Wong2008,Leader2007,Zheng2007} and normal biological variation\cite{Brown2017}. In most of the validation experiments, area measurements were taken from phantom\cite{Wong2008,Leader2007}, porcine\cite{Hammond2017, Leutz-Schmidt2017} \secondnewtext{or cadaver\cite{Zhang2019}} models. In Fetita et al.\cite{Fetita2014}, they used synthetic models of the lung. None of these experiments were explicitly performed on scans with bronchiectasis. Furthermore, the area measurements were not taken at contiguous intervals along the lumen thus missing possible dilatations from a bronchiectatic airway.

For our work and the method from Oguma et al.\cite{Oguma2015}, the tapering measurement involves the computation of the arc length of the airway at contiguous intervals. In the literature, investigations of the reproducibility of arc length computation in airways are limited. In Palagyi et al.\cite{Palagyi2006} they used simulated rotation of in vivo scans. The assessment of the reproducibility was based on the lengths of a single branch rather than multiple generations of branches, thereby precluding estimations of reproducibility of airway quantitation from the carina to an airway's most distal point.

\subsection{Contributions of the Paper}

To our knowledge, there is no detailed analysis on the reproducibility of a global tapering measurement of airways using CT. Thus, the purpose of this paper is as follows: first of all, we will discuss in detail a tapering measurement of the airways on CT imaging. Secondly, we quantify the reproducibility of the measurement against variations in simulated dose and voxel sizes. In addition, we compare the variability of the tapering measurements across different CT reconstructions \newtext{kernel}. Finally, we analyse the effect of bifurcations on tapering measurements, and consider measurement repeatability using longitudinal scans. 

\section{Method} \label{firstMethod}

\begin{figure}
	\centering
	\includegraphics[trim={0 0 0 0},clip,width=1\textwidth]{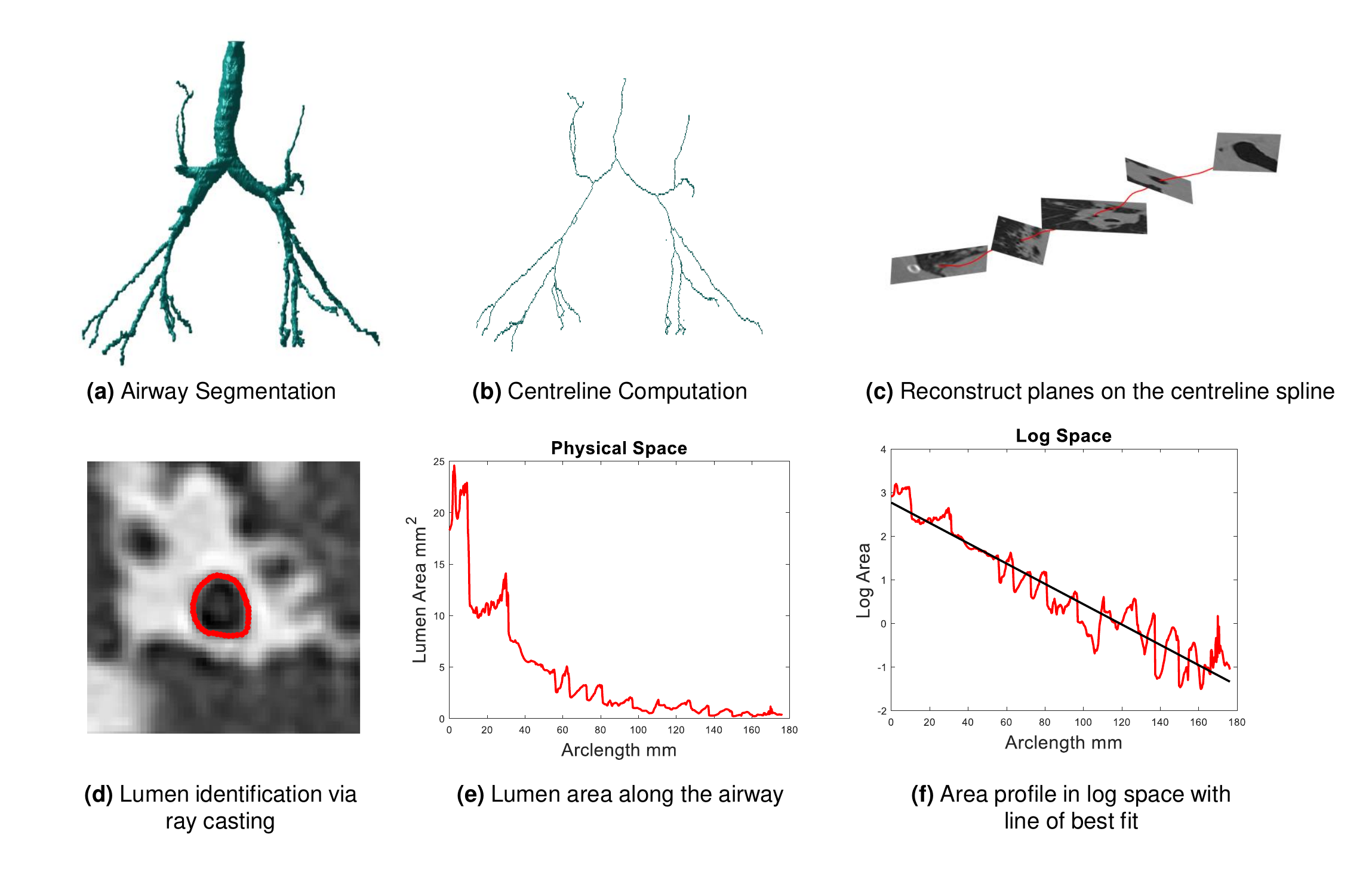}
	\caption{Summary of steps in our pipeline}
	\label{Pipeline_method}
\end{figure}

We first, describe in detail the steps to acquire the airway tapering measurement. The method was initially proposed by Quan et al.\cite{Quan2018} and summarised in Figure \ref{Pipeline_method}. The pipeline required two inputs. First, the most distal point of each airway of interest was manually identified by an experienced radiologist (JJ). A single voxel was marked at the end of the airway centreline. The entire analysis was completed using ITK-snap\footnote{\url{http://www.itksnap.org}, last accessed \today}.

Secondly, a complete segmentation of the airway was produced. We obtained an airway segmentation by implementing a method developed by Rikxoort et al.\cite{Rikxoort2009} The algorithm was based on a region growing paradigm. In summary, a wave front was initialised from the trachea. Voxels on each new iteration were classed as airways based on a voxel criterion. The wave front continued until a wave front criteria was met. In certain cases, the airway segmentation was unable to reach the distal points and in these cases we extended the airway segmentation manually to the distal points. Our method is designed so that it can be incorporated to any system that provides both the segmentation and distal point to the airway of interest. Once the inputs were available, we acquired the measurement using an automatic process.

\subsection{Centreline}

The centreline was used to identify and order the airway segments for the tapering measurement. We implemented a curve thinning algorithm developed by Palagyi et al.\cite{Palagyi2006} At initialisation, the algorithm used the airway segmentation and distal points acquired in Section \ref{firstMethod}. The final input was the start of the centreline at the trachea. The shape of the trachea was assumed to be tubular, with an approximate constant diameter and orientated near perpendicular to the axial slice. Thus, the centreline of the trachea lay on the local maximum value of the distance transform of the segmented trachea\cite{Aylward2002}. Algorithm \ref{Start_finder} was used to find the centreline start point.

\begin{algorithm}
	\SetAlgoLined
	\KwData{$D(i)$ Distance image on the $i$th axial slice}
	\KwResult{\replacetext{${x_{s}}$}{$\bm{x_{s}}$} Start point of trachea}
	$i\leftarrow $ First slice at the top of the trachea.\\
	\While{$\max{D(i)} < \max{D(i+1)}$}{
		$i = i + 1$
	}
	\replacetext{${x_{s}}$}{$\bm{x_{s}}$}$ = \argmax{D(i)}$
	\caption{Locating the start of centreline on the trachea}\label{Start_finder}
\end{algorithm}

\subsection{Recentring \& Spline fitting}
The next task was to separate the centreline of each individual airway from the centreline tree. To this end, we modelled the centreline tree as a graphical model similar to Mori et al.\cite{Mori2000} The nodes corresponded to the centreline voxels and the edges linked neighbouring voxels. We performed a breadth first search algorithm\cite{Cormen2009} on the centreline image. Starting from the carina, we iteratively found the next set of sibling branches. When a distal point was found at the end of a parent branch, the path leading to the distal point was saved. The output was an array of ordered paths describing the unique route from the trachea to the distal point. The proposed tapering measurement started at the carina. Thus, centreline points corresponding to the trachea were removed from further analysis.

For each path we corrected for the discretization error - a process known as re-centring\cite{Grelard2017}. We implemented a similar method to that described by Irving et al.\cite{Irving2014} A five point smoothing was performed along each path. We modelled the centreline as a continuous model by fitting a cubic spline $F:[0, k_{n}] \rightarrow \mathbb{R}^3$ denoted as

\begin{align}
\bm{F}(t)= 
\begin{cases}
\bm{f}_{1}(t),& t \in [0 , k_{1}]\\
~~\vdots              &  \\
\bm{f}_{n}(t) & t \in [k_{n-1} , k_{n} ]
\end{cases},
\end{align}
where $\bm{f}_{i}(t) = \sum_{j = 0}^{3} c_{i,j}t^{j}$ and $c_{i} \in \mathbb{R}^3$. \persontext{We exchanged  $f$ with $\bm{f}$ and $F$ with $\bm{F}$ in the equation above} The knots $k_{i}$ where taken on every smoothed point on the centreline. The spline fitting was performed using the \texttt{cscvn}\footnote{\url{https://uk.mathworks.com/help/curvefit/cscvn.html}, last assessed \today} function in Matlab. The continuous model should enable computations of the arc length and tangent at sub-voxel intervals along the airway.

\subsection{Arc Length}
The tapering measurement required an array of arc lengths at contiguous intervals from the carina to the distal point. For our pipeline, we considered small parametric intervals $t_{i}$ on the cubic spline \replacetext{$F(t)$}{$\bm{F}(t)$}. At each interval $t_{i}$, we computed the arc length from the carina to $t_{i}$ as\cite{Kreyszig1964}
\begin{equation}
s(t_{i}) = \int_{0}^{t_{i}} \sqrt{\dot{\bm{F}} \cdot \dot{\bm{F}}} ~ dt,
\end{equation}
where \newtext{$\dot{\bm{F}} = \frac{d\bm{F}}{dt}$ and } $(\cdot)$ is the dot product.  \persontext{We exchanged  $\frac{d\bm{F}}{dt}$ with $\dot{\bm{F}}$ in the equation above} For our work, we considered parametric intervals of 0.25 along the spline.

\subsection{Plane Cross Section}
We measured the cross-sectional area accurately by constructing a cross-sectional plane perpendicular to the airway. Using the interval $t_{i}$ from the arc length computation, we computed tangent vector \replacetext{$q$}{$\bm{q}$}$ \in \mathbb{R}^3$ by
\begin{equation}
\bm{q}(t_{i}) = \frac{\dot{\bm{F}}}{|\dot{\bm{F}}|},
\end{equation}
\persontext{We exchanged  $g$ with $\bm{g}$ and $F$ with $\bm{F}$}

From linear algebra, points on the plane can be generated by their corresponding basis vector\cite{Leon2009}. To this end, we generated a set of orthonormal vectors \replacetext{$v_{1}$}{$\bm{v}_{1}$}, \replacetext{$v_{2}$}{$\bm{v}_{2}$}$ \in \mathbb{R}^2$ using the method stated in Shirley and Marschner\cite{Shirley2009}. The method is summarised in Algorithm \ref{Normal_basis}.

\begin{algorithm}
	\SetAlgoLined
	\KwData{\replacetext{$q$}{$\bm{q}_{1}$} Unit tangent vector of the spline}
	\KwResult{ \replacetext{$v_{1}$,$v_{2}$}{$\bm{v}_{1},\bm{v}_{2}$} Basis of the orthogonal plane}
	\replacetext{$a$}{$\bm{a}$}$ \leftarrow $ Arbitrary vector such that \replacetext{$a$}{$\bm{a}$} and \replacetext{$q$}{$\bm{q}$} are not collinear \\
	\replacetext{$v_{1} = \frac{a \times t}{|a \times t|}$}{$\bm{v}_{1} = \frac{\bm{a} \times \bm{q}}{|\bm{a} \times \bm{q}|}$}  \\
	\replacetext{$v_{2} = v_{1} \times t$}{$\bm{v}_{2} = \bm{v}_{1} \times \bm{q}$}
	\caption{Constructing the basis for the plane reconstruction, adapted from Shirley and Marschner \cite{Shirley2009}.}\label{Normal_basis}
\end{algorithm}

Assuming \replacetext{$F$}{$\bm{F}$}$(t_{i})$ was the origin, each point \replacetext{$u$}{$\bm{u}$} $\in \mathbb{R}^3$ on the plane can be written as:
\begin{equation}
\bm{u} = \alpha_{1}\bm{v}_{1} + \alpha_{2}\bm{v}_{2}.
\end{equation}
\persontext{We set the vectors on the above equation} We selected the scalars $\alpha_{1}, \alpha_{2} \in \mathbb{R}$ such that the point spacing are 0.3mm isotropically.

\subsection{Lumen Cross Sectional Area} \label{recon_image_plane}
We calculated the cross sectional area using the Edge-Cued Segmentation-Limited \secondnewtext{Full} \replacetext{Half Width}{Width Half} Maximum (FWHM\textsubscript{ESL}), developed by Kiraly et al.\cite{Kiraly2005} The method is as follows: the cross-sectional planes were aligned on both the CT image and airway segmentation. The intensities of the plane were computed for both images using cubic interpolation. Fifty rays were cast out in a radial direction, from the centre of the plane. Each ray sampled the intensity of the two planes at a fifth of a pixel via linear interpolation. Thus, each ray produced two 1D images with the first from the binary plane $r_{b}$, and second from the CT plane $r_{c}$. We then applied Algorithm \ref{ray_pt} to find boundary point $l$.

\begin{algorithm}
	\SetAlgoLined
	\KwData{The rays: $r_{b}:[0,p] \rightarrow \mathbb{R}_{[0,1]}$, $r_{c}:[0,p] \rightarrow \mathbb{R}$ where $p$ is the length from the centre to the border of the plane.}
	\KwResult{The position of the lumen edge, $l$.}
	$s \leftarrow $ The first index of the ray such that $r_{b}(s)<0.5$ \\
	$I_{max} \leftarrow $ Local maximum intensity in $r_{c}$ nearest to s\\
	$x_{max} \leftarrow $ The index such that $r_{c}(x_{max}) = I_{max}$\\
	$I_{min} \leftarrow $ Minimum intensity in $r_{c}$ from $0$ to $x_{max}$ \\
	$x_{min} \leftarrow $ The index such that $r_{c}(x_{min}) = I_{min}$ \\
	$l \leftarrow $ The index such that $r_{c}(l) = (I_{max} + I_{min}) \times 0.5$ and $l \in[x_{min} , x_{max}]$
	\caption{Summary of the FWHM\textsubscript{ESL}, adapted from Kiraly et al.\cite{Kiraly2005} The purpose of the algorithm was to find the point of the ray which crossed the lumen.}\label{ray_pt}
\end{algorithm}

The final output of the \replacetext{FHWM}{FWHM}\textsubscript{ESL} was an array of 2D points corresponding to the edge of the lumen. Finally, we fitted an ellipse based on the least square principle. The method was developed and implemented in Matlab by Fitzgibbon et al. \cite{Fitzgibbon1996} We considered the cross sectional area as the area of the fitted ellipse.

\subsection{Tapering Measurement}
We assumed for a healthy airway that the cross-sectional area was modelled by an exponential decay along its centreline. It has been shown in human cadaver studies that the average cross section area in a branch reduces at an exponential rate at each generation\cite{Nikiforov1985}. The same observation has been noted in porcine models\cite{Azad2016}. Using the decay assumption, we modelled the relationship between the arc length and the cross-sectional area as
\begin{equation}
y = Tx + \log{A}, \label{exp_model}
\end{equation}
\persontext{We changed $y = A\exp{Tx}$ to $Tx + \log{A}$} where $x$ is the arc length of the spline, $T$ is the proposed tapering measurement, $y$ is the cross-sectional area and $A$ is an arbitrary constant.

\deletetext{We converted Equation} 
\deletetext{into a usable form by applying a log transform, $\log(\cdot)$ on the right hand side. Thus, we obtained the following linear relationship between arc length and area as:}  \persontext{We removed $\log{y} = Tx + \log{A}$}

\replacetext{For}{In terms of} implementation, \newtext{for each airway track,} we considered the array arc length and cross-sectional area computed for each individual airway. \replacetext{Using the exponential assumption, a}{A} logarithmic transform $\log(x)$ was applied only on the cross-sectional area array. We fitted a linear regression on the signal, the tapering measurement is defined as the gradient from the line of best fit.

\section{Evaluation}
An experienced radiologist (JJ) selected a total of 74 airways from 10 scans. The CT images were analysed from 9 patients with bronchiectasis after obtaining written informed consent at the Royal Free Hospital, London. The voxel size ranged from 0.63-0.80mm in plane and 0.80-1.5mm slice thickness. The airways were classified as healthy (n = 35) or bronchiectatic (n = 39) by the same radiologist. Details including the make and model of the scanner are provided in Table \ref{table_of_airways}. \deletetext{Bronchiectasis can affect any part of the lung and the affected region can vary in size}\cite{Milliron2015} From our dataset, many of the airways affected by bronchiectasis came from two patients. We used the same airways for the simulated low dose and voxel size experiments. A subset of the same airways were used for CT reconstruction \newtext{kernel} and bifurcation experiments.
\afterpage{
\begin{landscape}
\begin{table}
	\centering
	\begin{tabular}{ c | c | c | c |c }
		Patients & Bronchiectatic Airways & Healthy Airways & Scanner & Voxel Size (x,y,z) mm\\ \hline
		bx500 & 0 & 9 & Toshiba Aquilion ONE &0.67, 0.67, 1.50 \\
		bx503 & 15 & 5 & Toshiba Aquilion ONE & 0.64, 0.64, 1.50 \\
		bx504 & 0 & 8 & Toshiba Aquilion ONE & 0.78, 0.78, 1.50 \\
		bx505 & 5 & 0 & Toshiba Aquilion ONE& 0.75, 0.75, 0.80 \\
		bx507 & 0 & 4  & Toshiba Aquilion ONE & 0.63, 0.63, 1.50 \\
		bx508 & 1 & 0 & GEMD CT750 HD& 0.80, 0.80, 1.00 \\ 
		bx511 & 0 & 6 & Toshiba Aquilion ONE & 0.78, 0.78, 1.50 \\
		bx512 & 1 & 0 & GEMD CT750 HD & 0.69, 0.69, 1.00 \\
		bx513 & 1 & 3 & Toshiba Aquilion ONE & 0.73, 0.73, 1.50 \\
		bx515 & 16 & 0 & Toshiba Aquilion ONE & 0.78, 0.78, 1.50\\
	\end{tabular}
	\caption{List of CT images used for the experiment. The table includes the number of classified airways, scanner and voxel size. Abbreviation: GEMD - GE Medical Systems Discovery.}
	\label{table_of_airways}
\end{table}
\end{landscape}
}
\subsection{Simulated Images}
In this experiment, we simulated images with differing radiation dose and voxel size. The purpose was to analyse the reproducibility of the tapering measurement against various properties of the CT image. Furthermore, we varied the noise and voxel sizes at regular intervals. Thus we also analysed the sensitivity of the tapering measurement against the given parameters. Finally, we investigated the reproducibility of cross-sectional area and airway length measurements with changes in dose and voxel sizes respectively.

\subsubsection{Dose}
\begin{figure}
	\centering
	\includegraphics[trim={0 0 0 0},clip,width=1\textwidth]{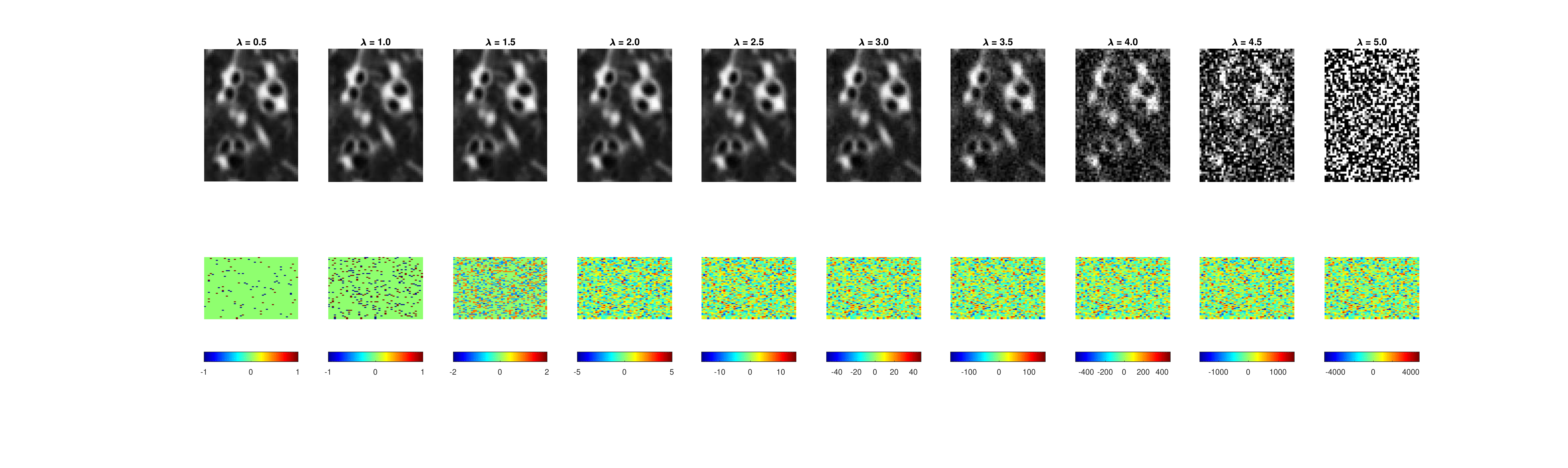}
	\caption{TOP: CT images with simulated noise against varying $\lambda$. BOTTOM: An image subtraction of the simulated noisy image with the original. The units are in HU.}
	\label{Dose_noise_method}
\end{figure}

To simulate the images acquired with different radiation doses, we used the method adapted from Frush et al.\cite{Frush2002} We performed a Radon transform on each axial slice of the original CT image. The output is a sinogram of the respective axial slice. To simulate different radiation doses, Gaussian noise was added on each sinogram with standard deviation $\sigma = 10^{\lambda}$; with a range of $\lambda$. The noisy sinograms were then transformed back into physical space using the filtered back projection. The final output is a noisy CT image in Hounsfield units in integer precision. A Matlab implementation is displayed on Algorithm \ref{mat_dose}. For our experiment we varied $\lambda$ from 0.5 to 5 in increments of 0.5. An example of the output image are displayed in Figure \ref{Dose_noise_method}.

\begin{algorithm}
	\SetAlgoLined
	function noisySlice = AddingDoseNoise(axialSlice,lambda)\\
	
	\%Creating the sinogram\\
	sinogram = radon(axialSlice,0:0.1:179);
	
	\%Adding the Gassian Noise \\
	noisySinogram = sinogram + randn(size(sinogram))*10\verb!^!(lambda);
	
	\%Converting the noisy sinogram into physical space \newtext{using Filter Backprojection.}\\
	noisySlice = iradon(noisySinogram,0:0.1:179,length(axialSlice));
	
	\%Converting into integer precision intensities\\
	noisySlice = int16(noisySlice);
	
	end
	\caption{Adapted Matlab code to simulate noise from differing doses}\label{mat_dose}
\end{algorithm}

To relate $\lambda$ to the physical dose from a CT scanner, we adopted the method described in Reeves et al.\cite{Reeves2017} This paper quantified the dose of an image with a homogeneous region in the chest CT scan. To this end, we used the homogeneous region inside the trachea. Using the airway segmentation, we considered the first 60 axial slices of the segmented trachea. To avoid the influence of the boundary, the tracheas were morphologically eroded\cite{Gonzalez2004} with a structuring element of a sphere of radius 5. All segmentations were visually inspected before further processing. Finally, we computed the standard deviation of the intensities inside the mask, denoted as $T_{n}$. Table \ref{table_of_noise}, shows values of $T_{n}$ on a selection of images against a range of $\lambda$. Using results from Reeves et al.\cite{Reeves2017} and Sui et al.\cite{Sui2016}, a low dose scan with a tube current-time product 25mAs has maximum $T_{n}$ of 55HU. Thus, we assume $\lambda = 3.5$ \secondnewtext{approximately} corresponds to a low dose scan. We considered higher values of $\lambda$ to verify any correlations in the results.

\afterpage{
\begin{landscape}
\begin{table}
	\centering
	\begin{tabular}{ c | c | c | c | c | c | c | c | c | c | c }
		$\lambda$ & bx500 & bx503 & bx504 & bx505 & bx507 & bx508 & bx511 & bx512 & bx513 & bx515 \\ \hline
		Ground Truth  & 15  & 28  & 15  & 21  & 16  & 28  & 20  & 33  & 21  & 17  \\ 0.5  & 15  & 28  & 15  & 21  & 16  & 28  & 20  & 33  & 21  & 17  \\ 1.0  & 15  & 28  & 15  & 21  & 16  & 28  & 20  & 33  & 21  & 17  \\ 1.5  & 15  & 28  & 15  & 21  & 16  & 28  & 20  & 33  & 21  & 17  \\ 2.0  & 15  & 28  & 15  & 21  & 16  & 28  & 20  & 33  & 21  & 17  \\ 2.5  & 16  & 29  & 16  & 22  & 17  & 28  & 20  & 33  & 22  & 17  \\ 3.0  & 21  & 32  & 21  & 26  & 22  & 31  & 25  & 36  & 26  & 22  \\ 3.5  & 49  & 54  & 49  & 51  & 49  & 54  & 51  & 57  & 51  & 49  \\ 4.0  & 146  & 150  & 148  & 148  & 148  & 148  & 149  & 149  & 148  & 145  \\ 4.5  & 461  & 467  & 467  & 463  & 466  & 462  & 467  & 459  & 463  & 457  \\ 5.0  & 1456  & 1473  & 1477  & 1464  & 1474  & 1458  & 1475  & 1449  & 1463  & 1445  \\ 
	\end{tabular}
	\caption{Table of standard deviation of intensity, $T_{n}$ (HU) in the inner lumen mask for a selected image against differing $\lambda$.}
	\label{table_of_noise}
\end{table}
\end{landscape}
}
We computed the taper measurement on the noisy images using the same segmented airways and labelled distal point that were identified on the respective original image. The literature has shown in low dose scans, airway segmentation software\cite{Rikxoort2009,Wiemker2006} cannot segment airways to the lung periphery as well as standard dose scans of the same patient. But these methods can still segment a large number of branches in low\cite{Rikxoort2009} and ultra-low\cite{Wiemker2006} dose scans. Furthermore, research has shown that there are minor differences in the performance of radiologists when attempting to detect features from standard and low dose CT scans\cite{Nagatani2015, Larbi2018, Sui2016}.

\subsubsection{Voxel Size}
We analysed the effect of voxel sizes on the tapering measurement. For each CT image, the voxel spacing $s_{x},s_{y},s_{z}$, was subsampled to new spacing of $\sigma s_{x}, \sigma s_{y}, \sigma s_{z}$, where $\sigma$ is a scalar constant. The intensities at each new voxel position was computed using sinc interpolation with a small amount of smoothing. We chosen Sinc interpolation to preserve as much information as possible from the original image. To compute the tapering value, we resampled the segmented airway and distal point to the same coordinate system using nearest neighbour interpolation. Morphological filtering via a closing operation\cite{Gonzalez2004} was used on segmented airways to remove artefacts caused by the resampling. For our experiment we used the parameters; $\sigma = 1.1, \ldots ,2$ with increments of $0.1$.

\subsection{CT Reconstruction}
\begin{figure}
	\centering
	\includegraphics[trim={0 0 0 0},clip,width=0.6\textwidth]{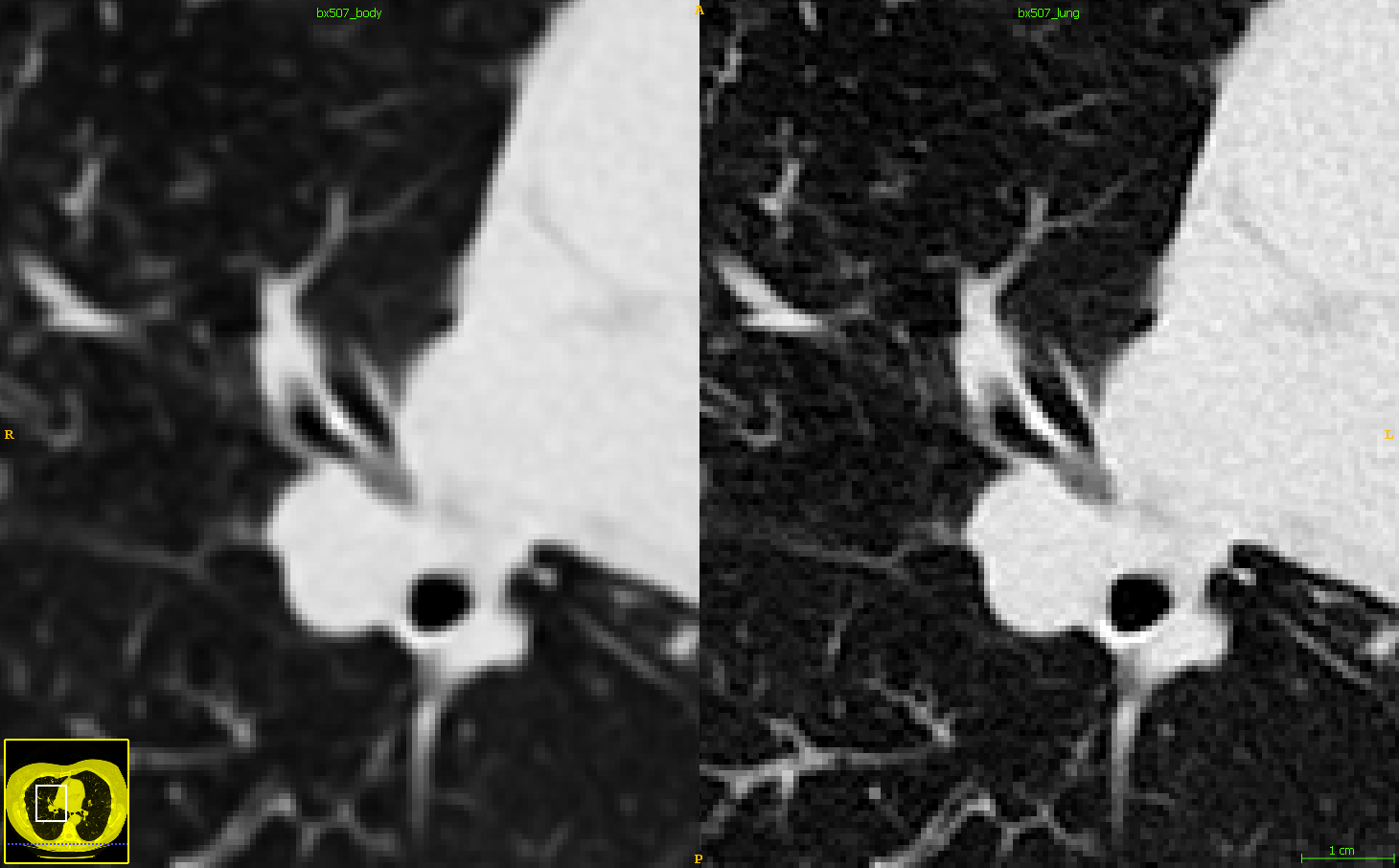}
	\caption{Images from the same CT scan with the body \replacetext{reconstruction}{kernel} (LEFT) and lung \replacetext{reconstruction}{kernel} (RIGHT). Both images are displayed in the same intensity window.}
	\label{body_and_lung_recon_bx507}
\end{figure}

On a subset of images, four patients were scanned using the Toshiba Aquilion One Scanner. On the same scan, two different images were computed. The images were reconstructed using the Lung and Body kernels respectively. An example of the reconstruction \newtext{kernels} is displayed on Figure \ref{body_and_lung_recon_bx507}. We acquired the airway segmentation and distal point from a single reconstruction \newtext{kernel} as described in Table \ref{recon_used}. The tapering measurement was computed on both reconstruction \newtext{kernels} using the same airway segmentation and distal points. We used the same airways as described in Table \ref{table_of_airways}. 

\begin{table}
	\centering
	\begin{tabular}{ c | c  }
		Patients & Reconstruction \newtext{Kernel} used for prepocessing  \\ \hline
		bx503 & Lung  \\
		bx507 & Body  \\
		bx513 & Lung  \\
		bx515 & Body \\
	\end{tabular}
	\caption{The images used for the reconstruction \newtext{kernel} experiment. The table lists which reconstruction \newtext{kernel} was used to generate the airways segmentation and distal point labelling. The make, model and voxels size of the images are displayed in Table \ref{table_of_airways}.}
	\label{recon_used}
\end{table}

\subsection{Biological Factors}

\subsubsection{Effect of Bifurcations} \label{Effect_of_Bifurcations}
We analysed the effect of airway bifurcations on the tapering measurement. To this end, we manually identified regions of bifurcating airways. On a selected subset of airways, we considered the reconstructed airway image described on Figure \ref{Bi_method_tapering}. Using ITK-snap, the author (KQ) started at the cross sectional plane corresponding to the carina and scrolled towards the distal point. Using visual inspection, the following protocol was developed to identify bifurcations on cross sectional planes:

\begin{figure}
	\centering
	\includegraphics[trim={0 0 0 0},clip,width=1\textwidth]{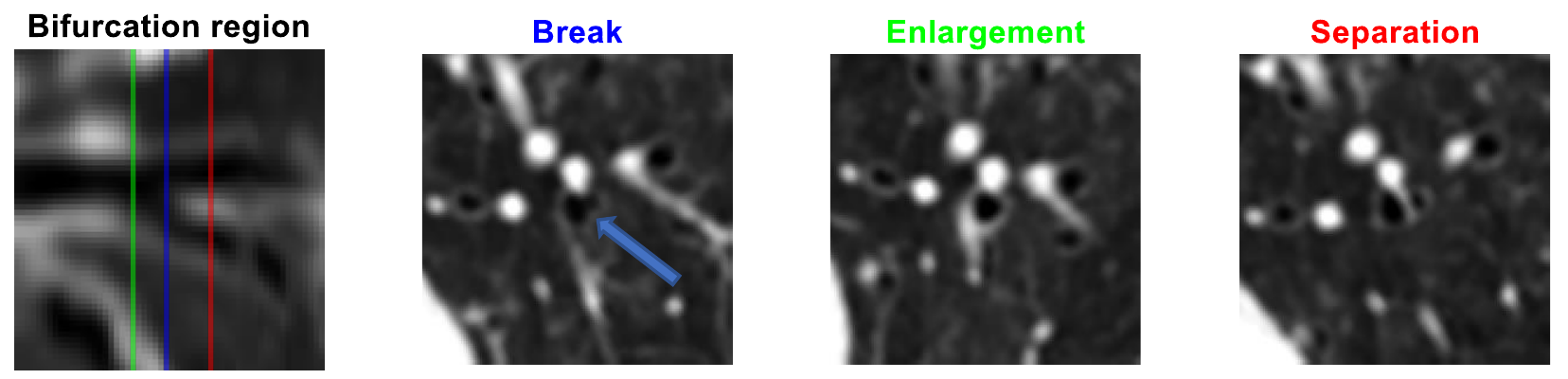}
	\caption{FAR LEFT: A region of bifurcation along the reconstructed slices. The green, blue and red regions are the slices corresponding to the enlargement, break and separation slices respectively. The labelled region consists of slices from green to red. CENTRE LEFT: A cross sectional plane where the airway is at the point of bifurcation, indicated by the blue arrow. CENTRE RIGHT: First slice of the bifurcation region. FAR RIGHT: The final slice of the bifurcation region. \newtext{The slides are chronologically ordered with the protocol described in Section \ref{Effect_of_Bifurcations}}}
	\label{Bi_method_with_arrow}
\end{figure}

\begin{enumerate}
	\item The scrolling stops when the airway is almost or at the point of separation.
	\item The author scrolls back until the airway stops decreasing in diameter. An alternative interpretation is when the airways are about to enlarge due to the bifurcation.
	\item Starting at the point of enlargement and scrolling forward, each slice is delineated as a bifurcating region until complete separation of bifurcating airways is reached. The criteria for a complete separation is the lumen wall of both airways are completely visible and separate. The entire protocol is summarised in Figure \ref{Bi_method_with_arrow}.
\end{enumerate}

For our experiment, we selected 19 airways from Table \ref{table_of_airways}. The data consisted of 11 healthy and 8 bronchiectatic airways. The entire analysis was performed on ITK snap.

\subsubsection{Progression}
We examined possible changes in tapering of airways in patients over time. \newtext{In this experiment, we consider two sets of longitudinal scans. First, pairs of airways that were healthy on both baseline and follow up scans. Secondly, pairs of airways that were healthy on baseline scans and became bronchiectatic on follow up scans.}

\begin{figure}
	\centering
	\includegraphics[trim={0 0 0 0},clip,width=1\textwidth]{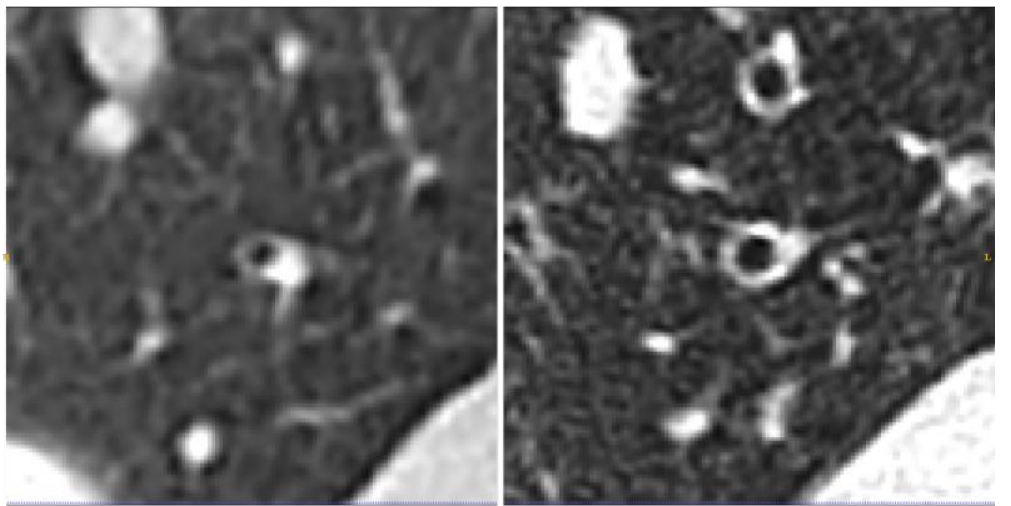}
	\caption{\persontext{New image} \newtext{The same pair of airways from longitudinal scans. LEFT: Initial healthy airway. RIGHT: The same airway at the same location becoming bronchiectatic.}}
	\label{H_to_D_airways}
\end{figure}

\newtext{For pairs of healthy airways,} \replacetext{A}{a} trained radiologist (JJ), manually identified 14 pairs of \deletetext{healthy} airways across 3 patients.   \deletetext{Three patients were chosen who had longitudinal scans,}. \newtext{The criteria were the airway track must have a healthy appearance on both baseline and follow up scans. For the second population, the same radiologist manually identified 5 pairs of airways from a single patient, P1. The scans were obtained from University College London Hospital and acquired with written consent. The criteria for selection were airways that appear healthy on baseline scans and became bronchiectatic at follow up scan, an example are displayed on Figure \ref{H_to_D_airways}. Details of the CT images are summarised on \ref{progression_info_table} and \ref{long_make_and_voxel_size}. }

Using two separate work stations, the airways were visually registered between the longitudinal scans. \newtext{Airways were taken from various regions of the lungs and were different to the airways displayed on Table} \ref{table_of_airways}. \deletetext{The chosen airways were separate form Table} \ref{table_of_airways}. The tapering measurements were taken from the method discussed in Section \ref{firstMethod}.

\begin{landscape}
\begin{table}
	\centering
	\begin{tabular}{ c  | c | c }
		Patients & Time between scans & Airways Labelled \\ \hline
		bx500 & 9M 6D & 6 \\
		bx504 & 35M 6D & 7 \\
		bx510 &  5M 22D & 1 \\ \hline
		\newtext{P1} & \newtext{10M 7D} & \newtext{5}
	\end{tabular}
	\caption{List of the images for progression experiment. The table includes time between scans in months (M) and days (D). \newtext{The airways on this table are different to Table \ref{table_of_airways}}}
	\label{progression_info_table}
\end{table}

\begin{table}
	\centering
	\begin{tabular}{ c | c | c | c | c }
		Patients & Date 1 CT Scanner & Date 1 Voxel Size & Date 2 CT Scanner & Date 2 Voxel Size \\ \hline
		bx500 & Toshiba Aquilion One & 0.67, 0.67, 1.00 & Toshiba Aquilion One & 0.56, 0.56, 1.00 \\
		bx504 & Toshiba Aquilion One & 0.63, 0.63, 1.00 & Toshiba Aquilion One & 0.78, 0.78, 1.00 \\
		bx510 & GEMS LightSpeed Plus & 0.70, 0.70, 1.00 & Philips Brilliance 64 & 0.72, 0.72, 1.00 \\ \hline
		\newtext{P1} & \newtext{GEMS Discovery STE} & \newtext{0.85, 0.85, 2.50} & \newtext{SS Definition AS} & \newtext{0.66,0.66,1.50}
	\end{tabular}
	\caption{List of make, models and voxel sizes of CT images for progression experiment. The voxel sizes are displayed as x,y,z and in mm units.  Abbreviation: GEMS - GE Medical Systems, \newtext{SS - Siemens SOMATOM}.}
	\label{long_make_and_voxel_size}
\end{table}
\end{landscape}

\section{Results}
\begin{figure}
	\centering
	\includegraphics[trim={0 0 0 0},clip,width=0.8\textwidth]{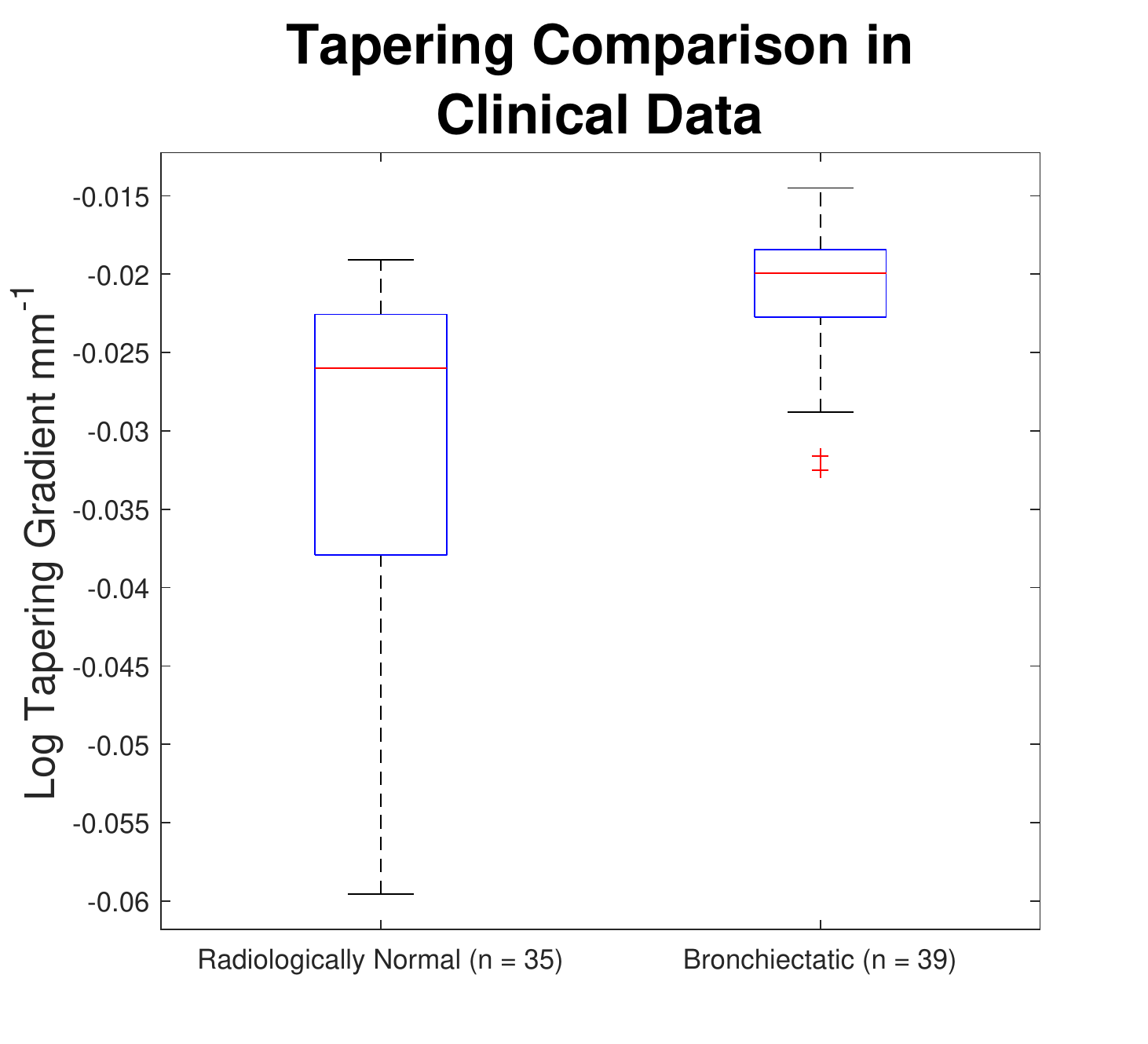}
	\caption{Comparing the proposed tapering measurement with labelled healthy and bronchiectatic airways. On a Wilcoxon Rank Sum Test between the populations, $p = 3.4 \times 10^{-4}$.}
	\label{Results_tapering_comparision}
\end{figure}

Figure \ref{Results_tapering_comparision} compares the tapering measurement between healthy and diseased airways. On a Wilcoxon Rank Sum Test between the populations, $p = 3.4 \times 10^{-4}$.

\subsection{Dose}
We analysed the difference in cross-sectional area measurements and the final tapering measurements at different CT radiation doses.\\

\begin{sidewaysfigure}
	\centering
	\includegraphics[trim={0 0 0 0},clip,width=1\textwidth]{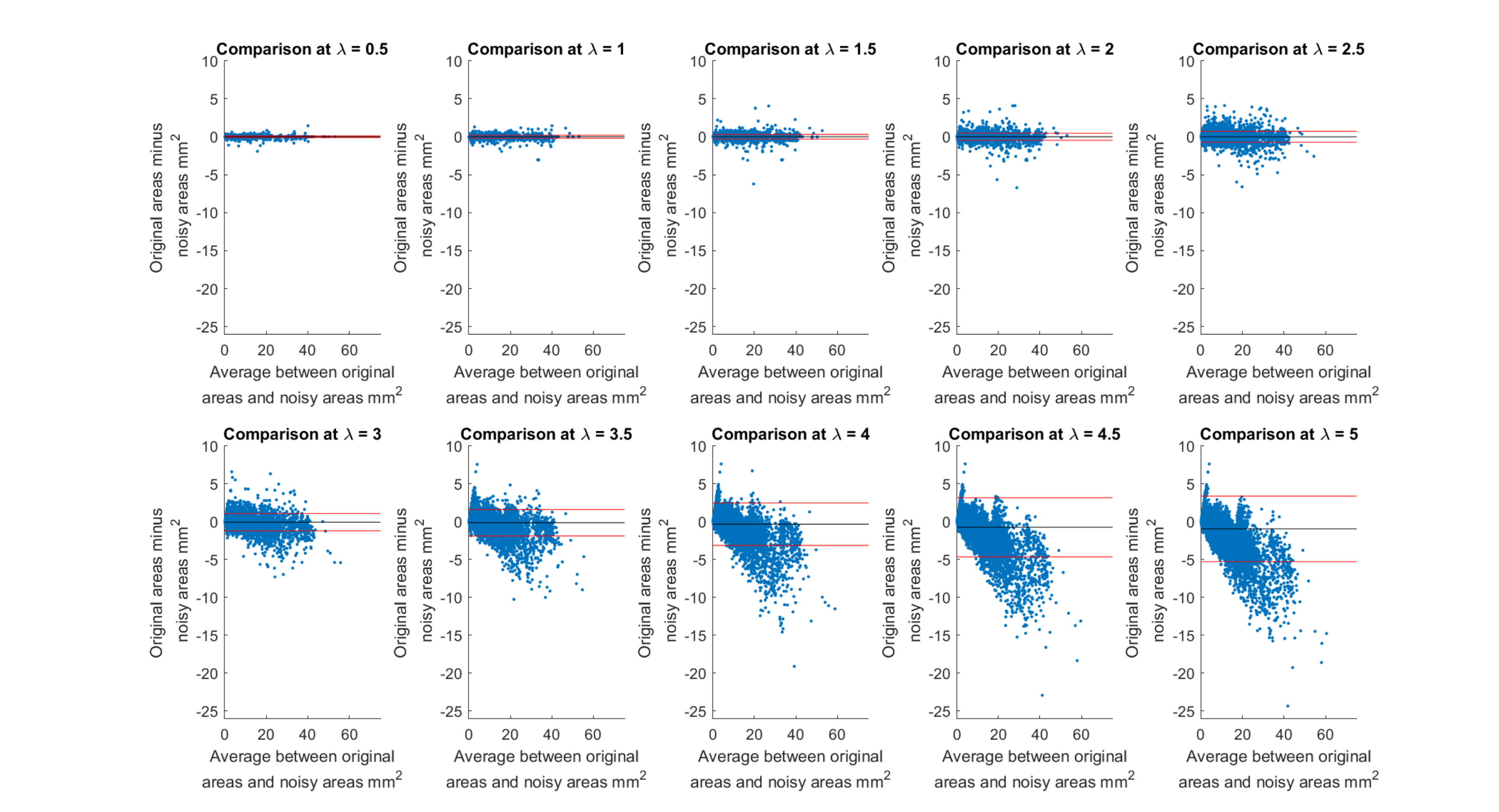}
	\caption{A series of Bland-Altman\cite{Bland1986} graphs comparing area measurements from a simulated low dose scan and the original image. On all graphs, the correlation coefficient was $r>0.99$}
	\label{Dose_noise_area}
\end{sidewaysfigure}

\begin{sidewaysfigure}
	\includegraphics[trim={0 0 0 0},clip,width=1\textwidth]{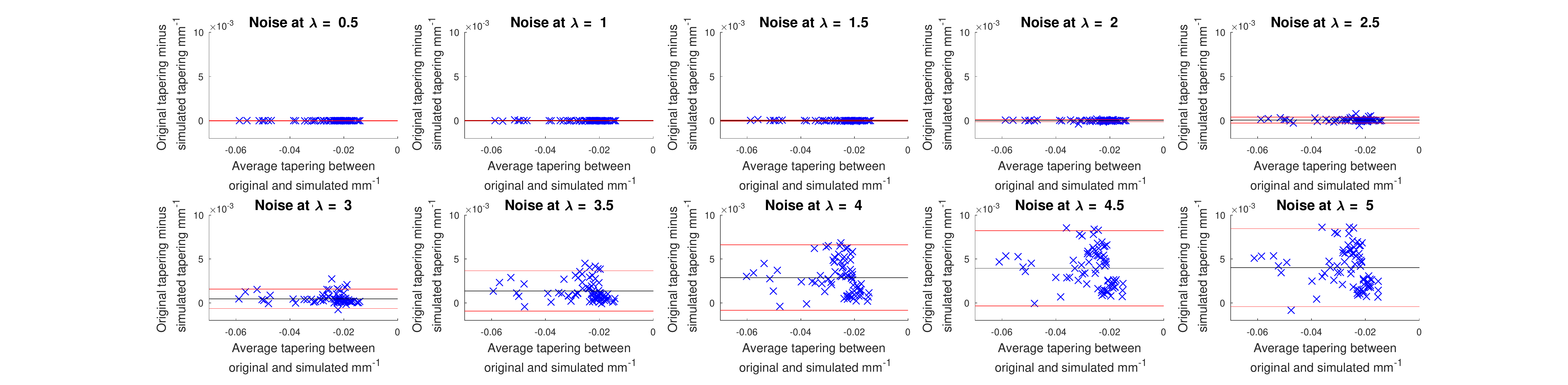}
	\caption{A series of Bland-Altman\cite{Bland1986} graphs comparing tapering measurement between simulated dose and the original image. On all graphs the correlation coefficient was $r>0.98$.}
	\label{Dose_noise_lamdba}
\end{sidewaysfigure}

For the cross-sectional areas, Figure \ref{Dose_noise_area}, compares the cross-sectional areas between the original image and one of the noisy images. Each graph contains approximately 30000 unique lumen measurements. The correlation coefficients between the populations was $r>0.99$ on all graphs. The 95\% confidence intervals increase with the amount of noise. For the tapering measurement, Figure \ref{Dose_noise_lamdba} displays the measurements from all the noisy images compared to their respective original images. The correlation coefficient between noisy and original tapering measurements was $r > 0.98$ on all values of $\lambda$.\\

\begin{figure}
	\centering
	\includegraphics[trim={0 0 0 0},clip,width=1\textwidth]{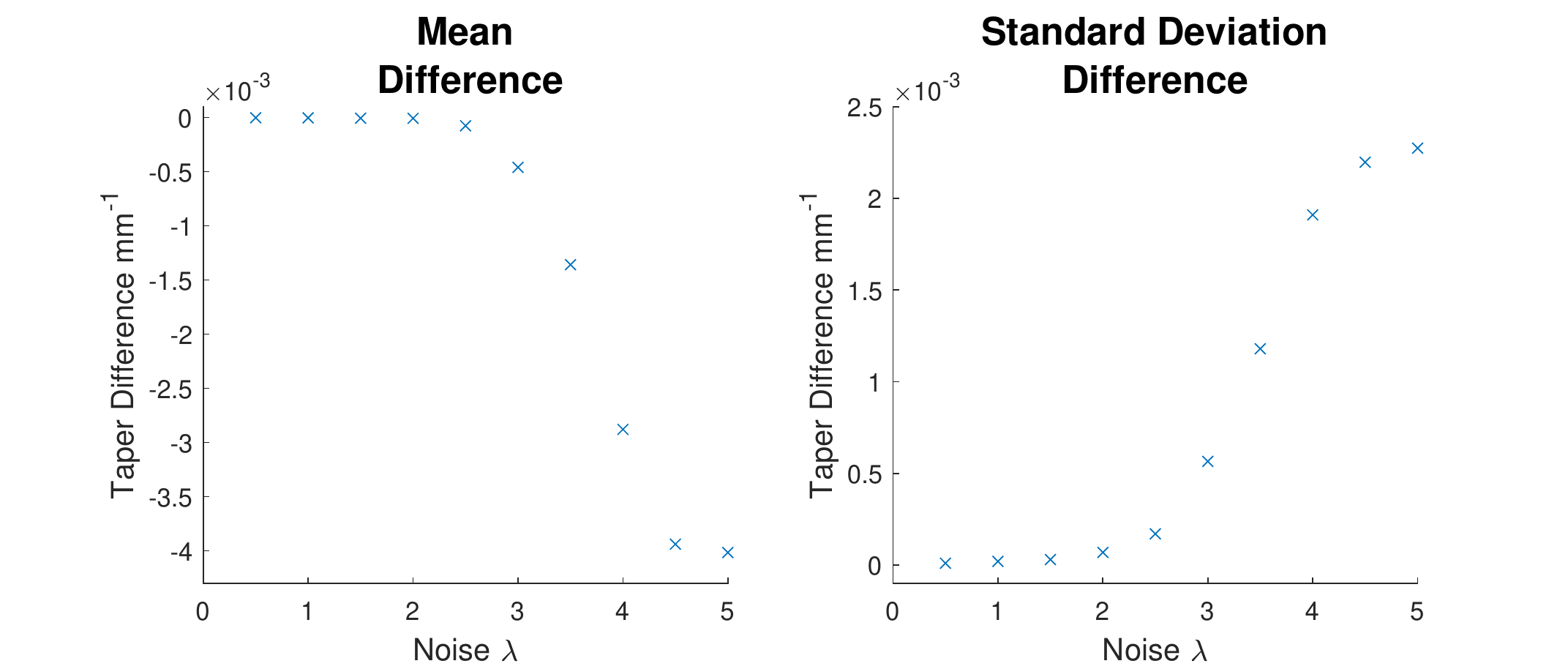}
	\caption{\persontext{Removed Error on the titles of the figures} Mean (LEFT) and standard deviation (RIGHT) of the difference in tapering between original images minus the simulated lower dose.}
	\label{Dose_noise_diff}
\end{figure}

We analysed the tapering difference between the original images and simulated images. We interpret the mean and standard deviation of the error difference as the bias and uncertainty respectively. Figure \ref{Dose_noise_diff}, shows an overestimation bias with an increase in noise and a positive correlation between uncertainty and dose.

\subsection{Voxel Size}
\begin{sidewaysfigure}
	\centering
	\includegraphics[trim={0 0 0 0},clip,width=1\textwidth]{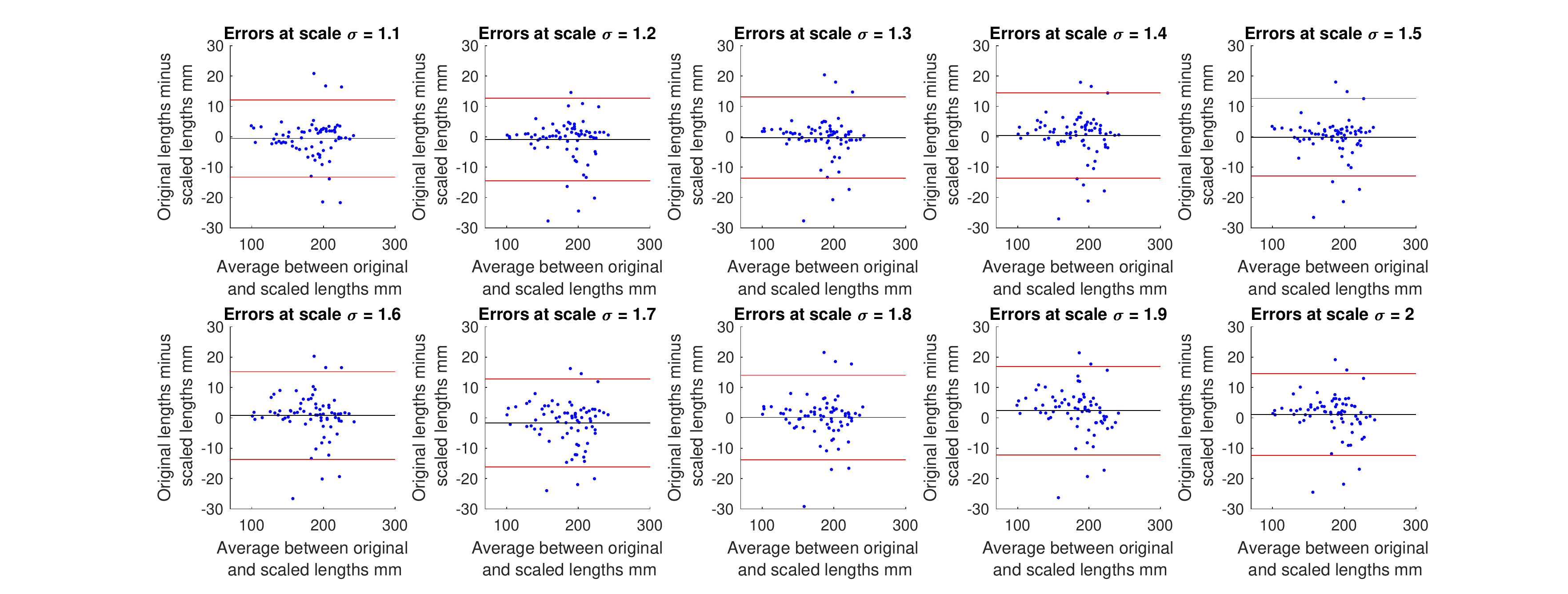}
	\caption{A series of Bland-Altman\cite{Bland1986} graphs comparing arc lengths between scaled images and the original images. On all graphs, the correlation coefficient was $r>0.98$}
	\label{Voxel_BA_lengths}
\end{sidewaysfigure}

\begin{figure}
	\centering
	\includegraphics[trim={0 0 0 0},clip,width=1\textwidth]{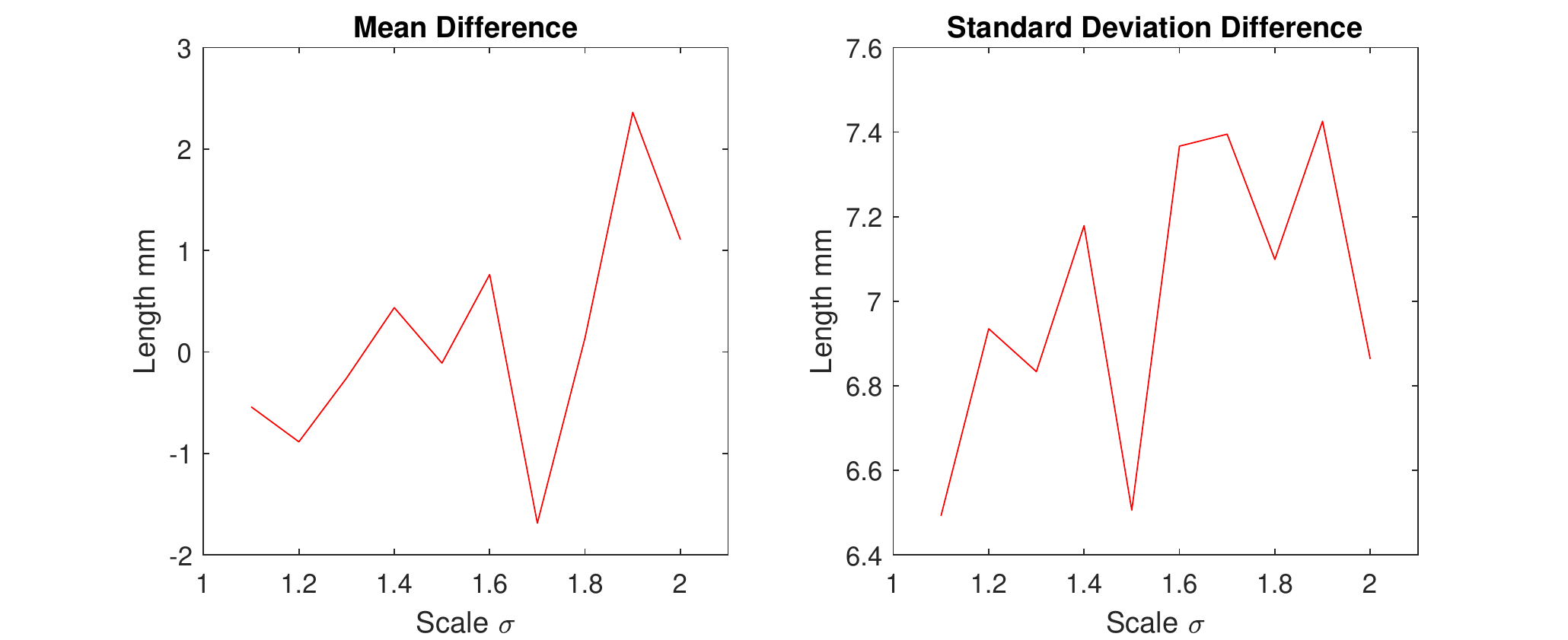}
	\caption{\persontext{Removed Error on the titles of the figures} Mean (LEFT) and standard deviation (RIGHT) of the difference in arclength between original images minus the scaled images. The correlation coefficient of the mean and standard deviation against scale are $r = 0.55$ and $r = 0.51$ respectively.}
	\label{Voxel_length_error}
\end{figure}

We analysed the computed spline and tapering for all the scaled images. We used the arclength of the spline as the metric for comparison for the computed spline. Figure \ref{Voxel_BA_lengths} compares the arclengths computed from the scaled splines with the respective originals. On all scales $\sigma$ the correlation coefficients between measurements was $r > 0.98$. Furthermore, we analysed the error difference in arclength. On Figure \ref{Voxel_length_error}, the mean difference shows a weak correlation coefficient with $r = 0.55$ with scale $\sigma$. The mean difference shows both an overestimation and underestimation bias with the arclength measurement. Figure \ref{Voxel_length_error}, shows a weak correlation between standard deviation and scale with $r = 0.51$.

\begin{sidewaysfigure}
	\centering
	\includegraphics[trim={0 0 0 0},clip,width=1\textwidth]{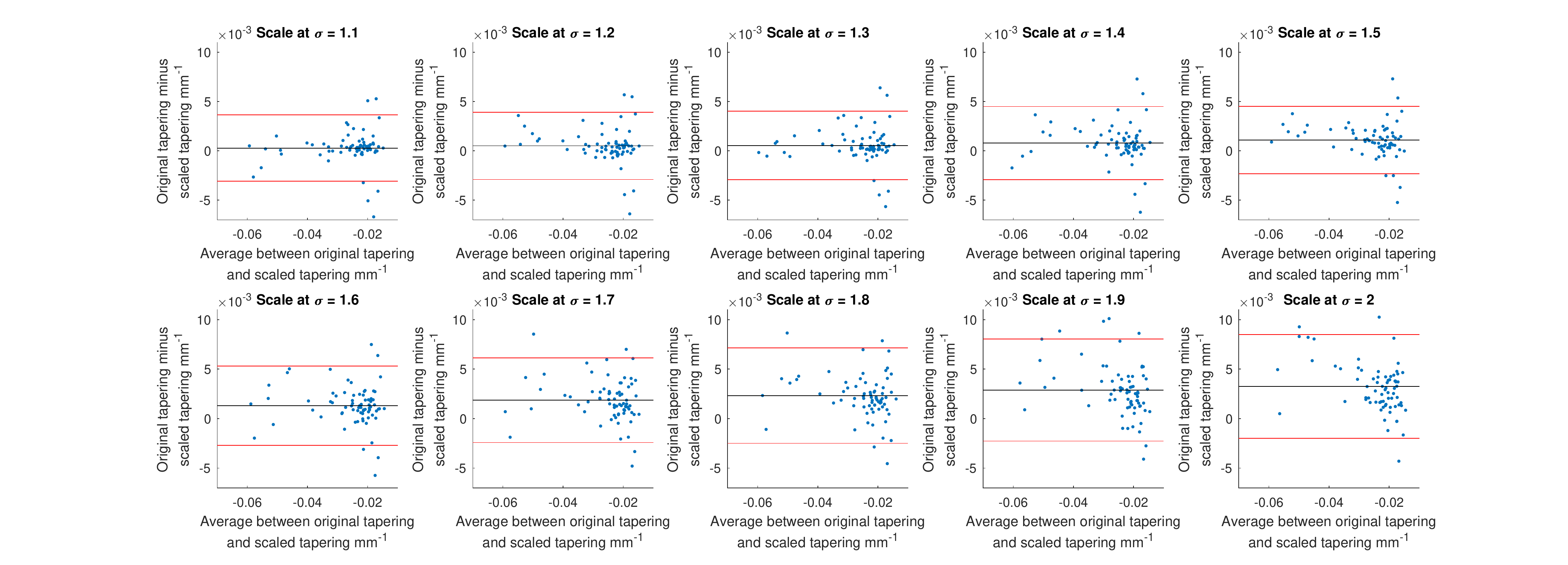}
	\caption{A series of Bland-Altman\cite{Bland1986} graphs comparing tapering between original images and scaled images. On all graphs the correlation coefficient was $r>0.97$}
	\label{Voxel_BA_diagram}
\end{sidewaysfigure}

\begin{figure}
	\centering
	\includegraphics[trim={0 0 0 0},clip,width=1\textwidth]{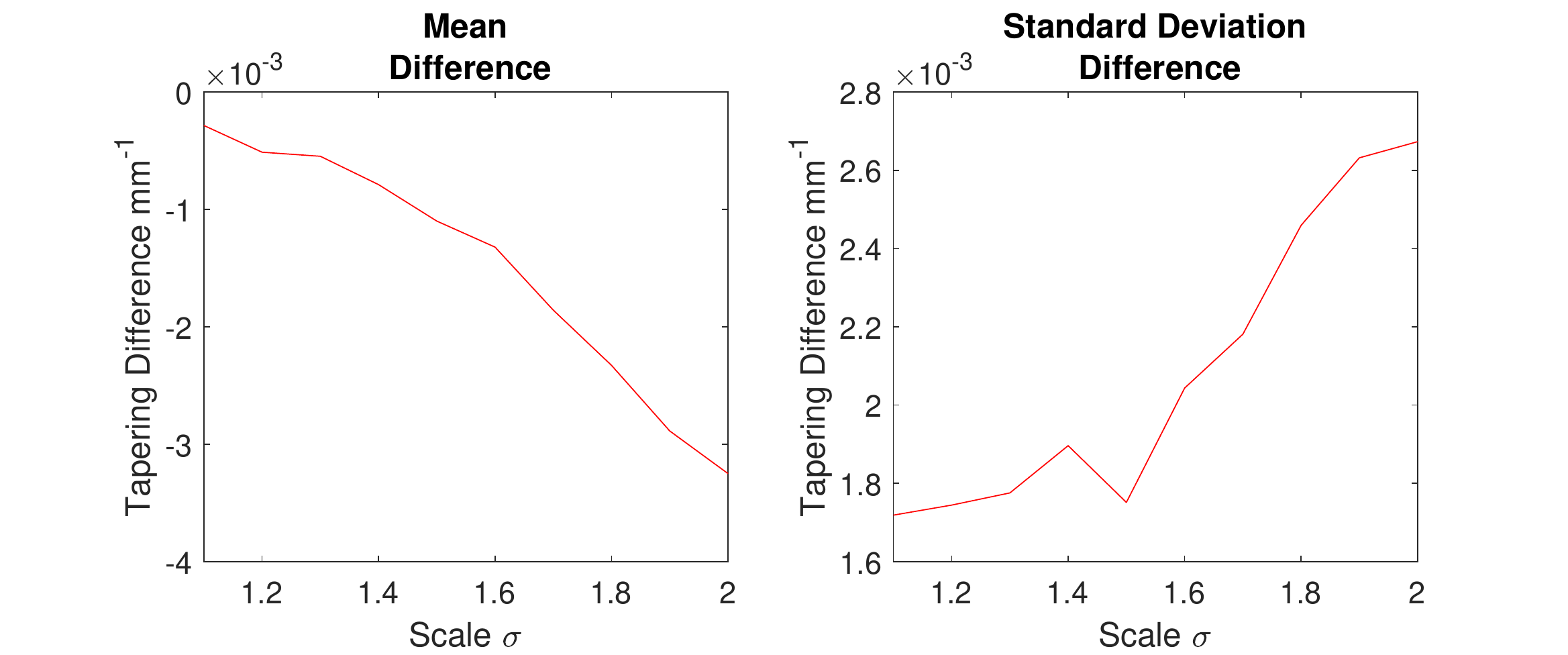}
	\caption{\persontext{Removed Error on the titles of the figures} Mean (LEFT) and standard deviation (RIGHT) of the difference in tapering between original images minus the scaled images. The correlation coefficient of the mean and standard deviation against scale are $r = -0.98 $ and $r = 0.94$ respectively.}
	\label{Voxel_taper_error}
\end{figure}

In terms of the tapering measurement, Figure \ref{Voxel_BA_diagram} compares the tapering values from the scaled images with the respective originals. The correlation coefficients between the scaled and original tapering values was $r > 0.97$ on all scales $\sigma$. In addition, we examined the error difference of the original minus the scaled tapering. Figure \ref{Voxel_taper_error}, shows a negative correlation with both overestimation and scale with $r = -0.98$. Furthermore, Figure \ref{Voxel_taper_error}, shows a positive correlation with uncertainty and scale with $r = 0.94$.

\subsection{CT Reconstruction}

\begin{figure}
	\centering
	\includegraphics[trim={0 0 0 0},clip,width=1\textwidth]{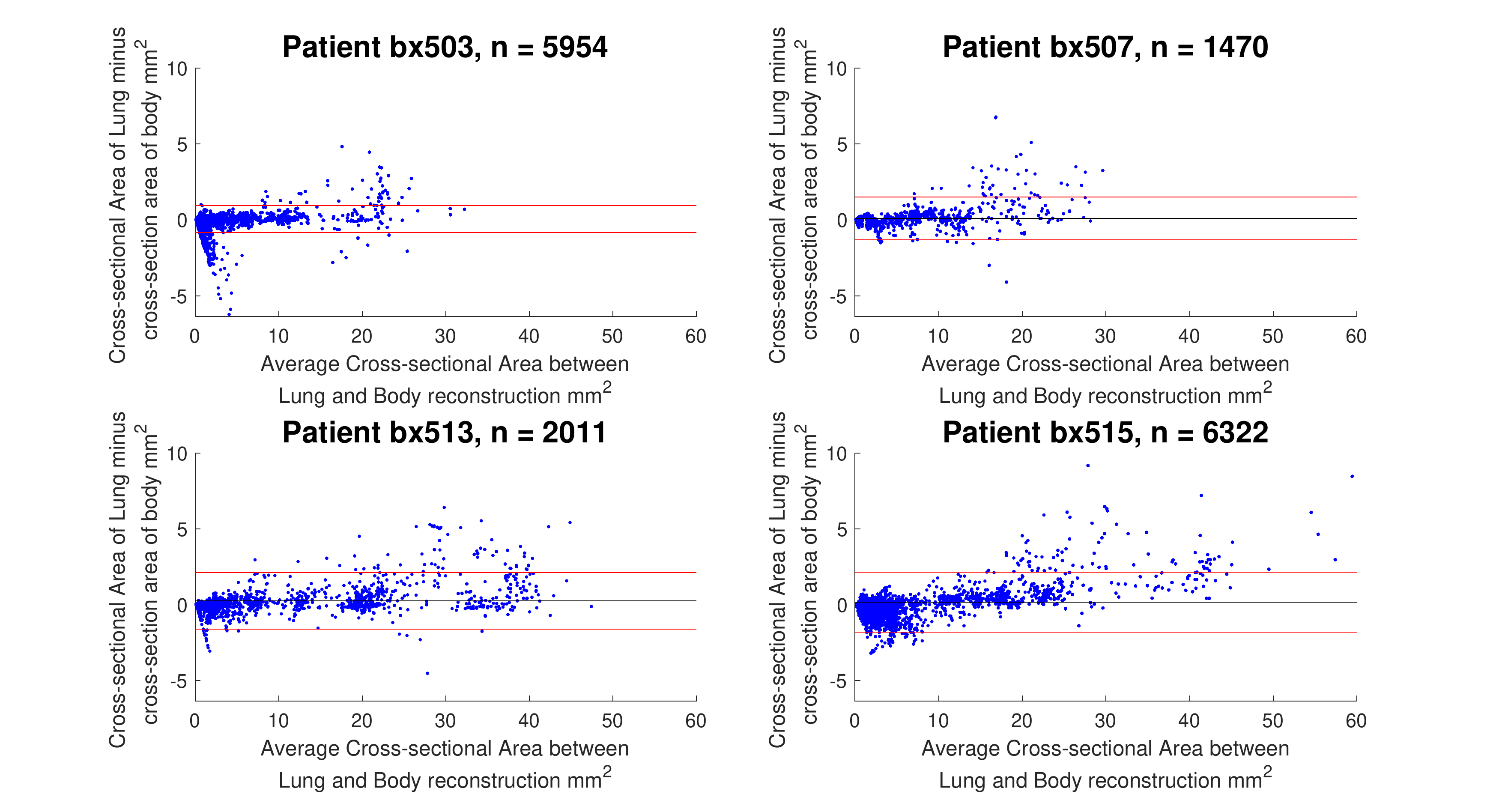}
	\caption{Bland-Altman\cite{Bland1986} graphs comparing the cross-sectional area between the Lung and Body \replacetext{reconstructions}{kernels}. On all four images the correlation coefficient was $r > 0.99$}
	\label{Recon_pat_area}
\end{figure}

\begin{figure}
	\centering
	\includegraphics[trim={0 0 0 0},clip,width=1\textwidth]{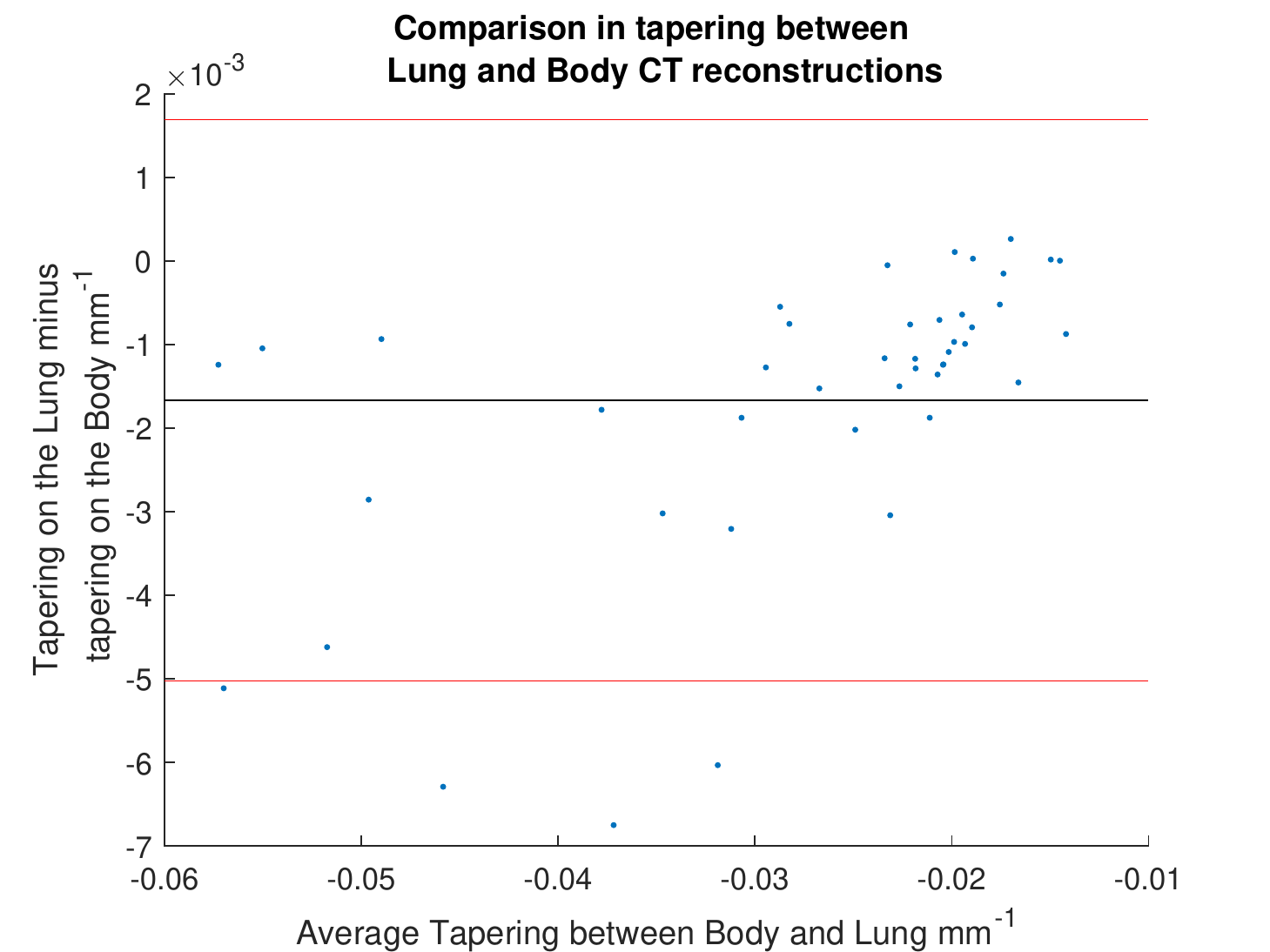}
	\caption{Bland-Altman\cite{Bland1986} graph comparing tapering measurements $(n=44)$ between Lung and Body  \replacetext{reconstructions}{kernels}, $r = 0.99$.}
	\label{Recon_pat_tap}
\end{figure}

We analysed the difference in cross sectional area and tapering measurement between reconstruction \replacetext{algorithm}{kernel}. Figure \ref{Recon_pat_area}, compares the difference in area measurements. On all patients, in cross sectional area measurements, the correlation coefficient between the two measurements was $r>0.99$. The largest 95\% confidence was in patient bx515 with $\pm 1.98$ mm$^2$ from the mean. Figure \ref{Recon_pat_tap}, compares the differences in tapering measurement. We collected $n = 44$ tapering measurement from 4 patients. The correlation coefficient was $r = 0.99$ between the reconstruction \newtext{kernel}.

\subsection{Clinical Results}

\subsubsection{Bifurcations}

\begin{figure}
	\centering
	\includegraphics[trim={0 0 0 0},clip,width=1\textwidth]{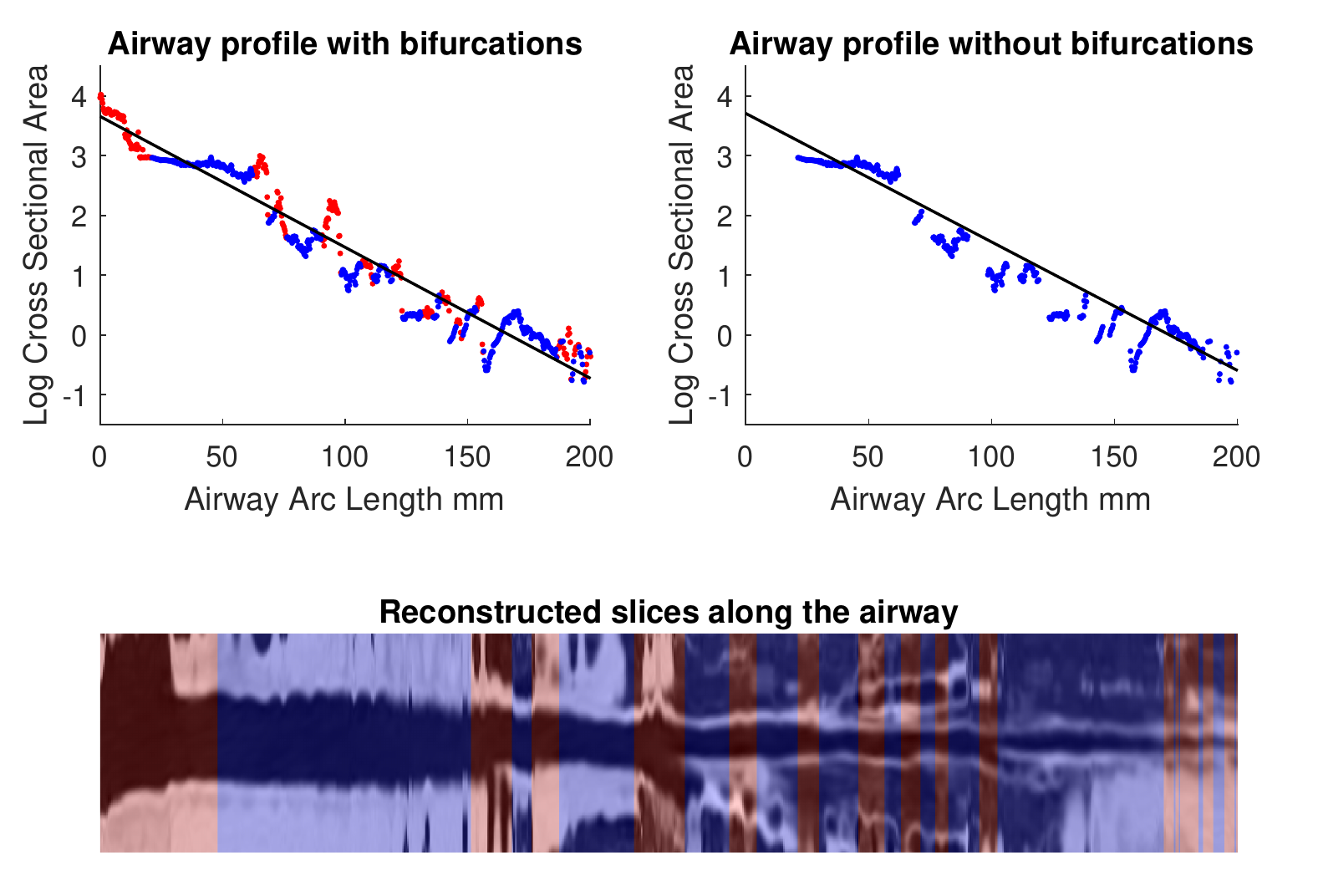}
	\caption{TOP LEFT: A signal of area measurement with bifurcation regions (red) and tubular regions (blue). TOP RIGHT: The same signal with tubular regions (blue) only. On both graphs, the black line is the linear regression of the respective data. The gradient of the line is the proposed tapering measurement. BOTTOM: The reconstructed bronchiectatic airway of the same profile. The blue-shaded and red-shaded regions corresponds to the tubular and bifurcating airways respectively. A reconstructed healthy airway have been discussed in Quan et al.\cite{Quan2018} Similar reconstructed cross sectional images of vessels have been discussed in Oguma et al.\cite{Oguma2015}, Kumar et al.\cite{Kumar2015}, Alverez et al.\cite{Alvarez2017}. and \secondnewtext{Kirby et al.}\cite{Kirby2018}}
	\label{Bi_method_tapering}
\end{figure}

\begin{figure}
	\centering
	\includegraphics[trim={0 0 0 0},clip,width=1\textwidth]{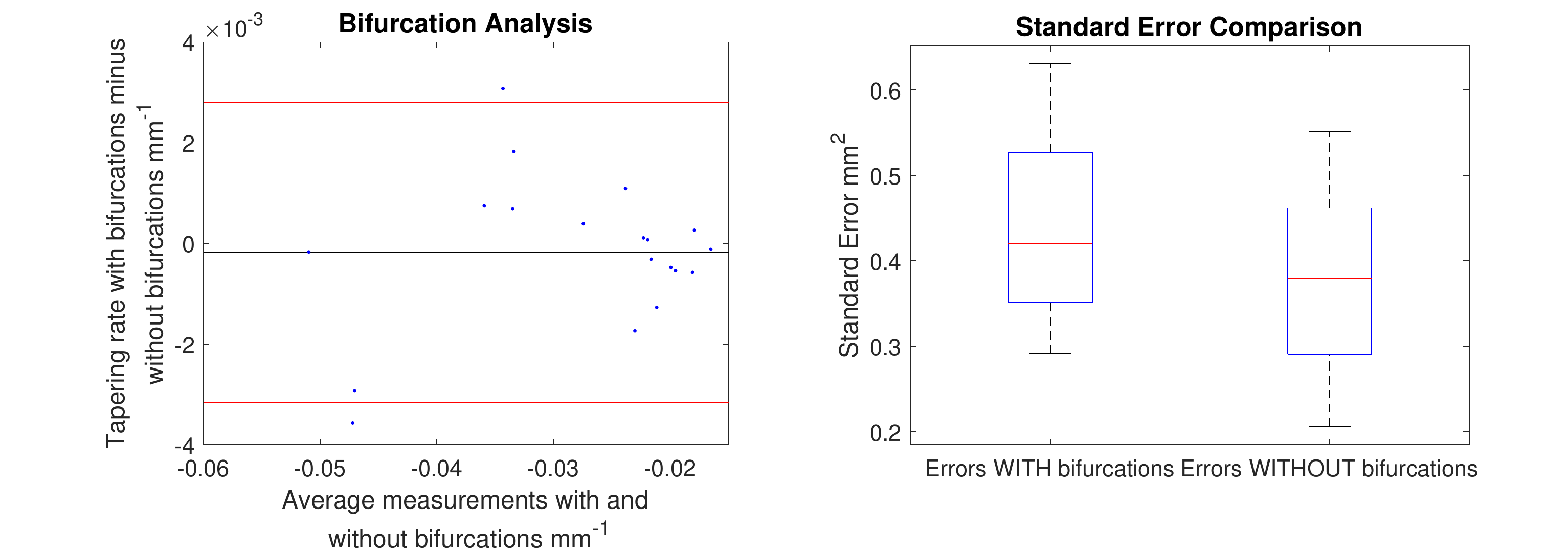}
	\caption{LEFT: Bland-Altman\cite{Bland1986} graph showing the relationship of the taper rates $(n = 19)$ with and without bifurcations, $r = 0.99$. RIGHT: Comparison of the standard error from linear regression between airways with and without bifurcations. On a Wilcoxon Rank Sum Test between the two populations, $p = 7.1 \times 10^{-7}$.}
	\label{Bi_results}
\end{figure}

We compared tapering measurements with and without points corresponding to bifurcations. On the first dataset, the tapering measurements were computed using all area measurements. The second dataset had tapering measurements computed without area measurements from the bifurcating regions as described in Figure \ref{Bi_method_tapering}. We compared the measurements on Figure \ref{Bi_results}, the correlation coefficient was $r = 0.99$.

The uncertainty of each tapering measurement was computed using the standard error of estimate $s$, defined as\cite{Spiegal1998}:
\begin{equation}
s = \sqrt{ \frac{\sum_{i = 1}^{N} (Y_{i} - y_{i})^2}{N}}
\end{equation}
where \deletetext{$x_{i}$},$y_{i}$ is the \deletetext{computed area and} arclength \deletetext{respectively}, $Y_{i}$ is the estimate from the linear regression from each \newtext{computed area} $x_{i}$ and $N$ is the number of points in the profile. Figure \ref{Bi_method_tapering} compares the uncertainty between the two populations. There was a statistical difference between the populations, on a Wilcoxon Rank Sum Test,  $p = 7.1\times10^{-7}$.

\subsubsection{Progression}
\begin{figure}
	\centering
	\includegraphics[trim={0 0 0 0},clip,width=1\textwidth]{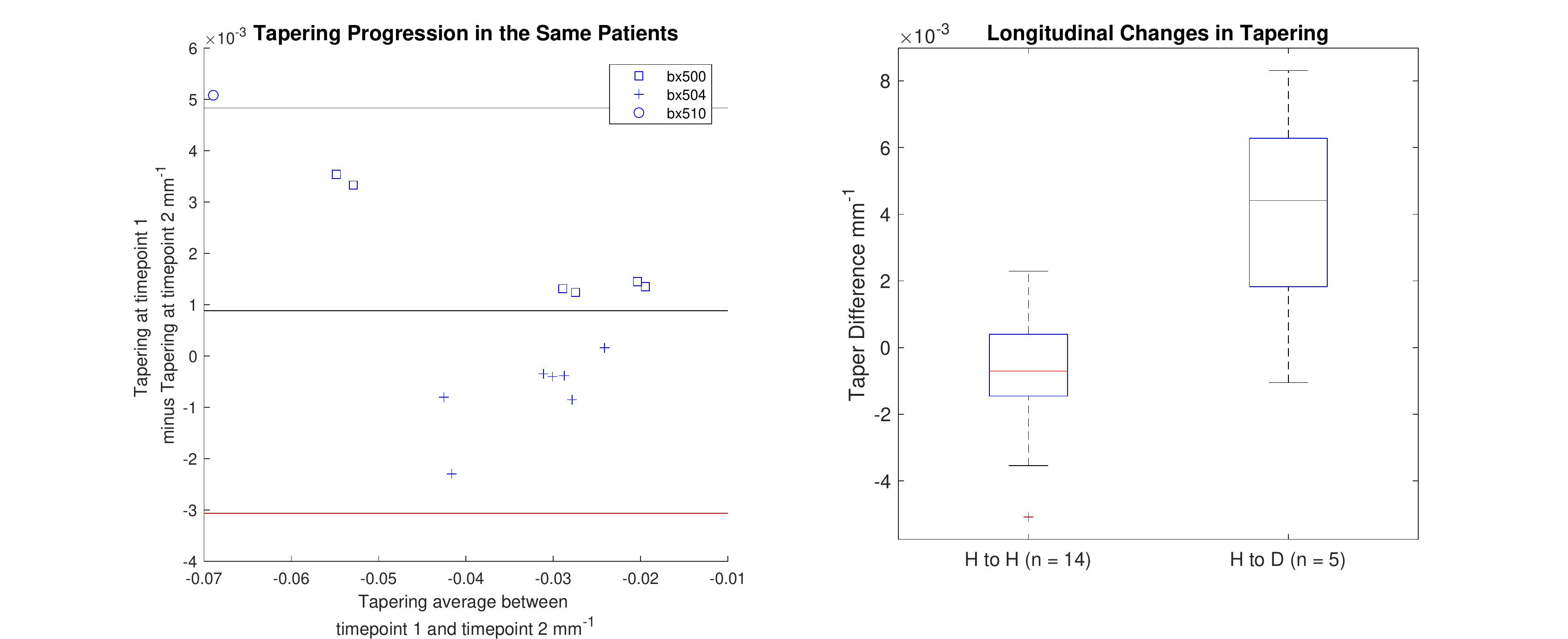}
	\caption{\persontext{Added more results} \newtext{RIGHT:} Bland-Altman\cite{Bland1986} graph comparing tapering measurement on the same airways from the first and subsequent scan, $ICC > 0.99$. \newtext{LEFT: The change in tapering between healthy airways (H to H) and airways that became bronchiectatic on the follow up scans (H to D), $p = 0.0072$.}}
	\label{Progression_taper_results}
\end{figure}

\newtext{For healthy airways,} \replacetext{W}{w}e grouped tapering values between the baseline and follow up point. Figure \ref{Progression_taper_results}, compares the measurements between the two \replacetext{populations}{time points}. The results demonstrated good agreement with an intraclass correlation coefficient\cite{Streiner2015} $ICC > 0.99$. The standard deviation of the tapering difference was $1.45 \times 10^{-3}$mm$^{-1}$.

\newtext{For airways that became bronchiectatic, we considered the change in tapering i.e. tapering value at follow up minus tapering value at baseline, the results are displayed on Figure \ref{Progression_taper_results}. The results shows bronchiectatic airways have a greater tapering change in magnitude compared to airways that remained healthy, $p = 0.0072$.}

\section{Discussions}
In this paper, we propose a tapering measurement for airways imaged using CT and validate the reproducibility of the measurement. The tapering measurement is the exponential decay constant between cross-sectional area and arclength from the carina to the distal point of the airway. Unlike other proposed tapering measurements, we assess reproducibility of the tapering measurement against simulated CT dose, voxel size and CT reconstruction \replacetext{algorithm}{kernel}. Finally, we assess the effect of tapering across airway bifurcations, and examine repeatability over time using longitudinal scans.

\newtext{Part of the evaluations consist of analysing the difference in tapering across longitudinal scans. The timescales between scans ranges from 9 to 35 months. The motivation for a long timescale is a proof of principle demonstration that the tapering measurement is reproducible for clinical studies. Examples include, drug trails\cite{DeSoyza2018} and investigations in exacerbations\cite{Chalmers2018a}, where the timescales in monitoring patients were 12 months and 60 months respectively.}

The pipeline consists of various established image processing algorithms. We chose the centreline algorithm developed by Palagyi et al.\cite{Palagyi2006}. Unlike other proposed methods\cite{Aylward2002,Jin2016,Cardenes2010} the algorithm explicitly links the distal points to the carina. Furthermore, it has been shown that the algorithm of Palagyi et al.\cite{Palagyi2006} can be used on images with non-isotropic voxel sizes\cite{Irving2014}. By modelling the centreline as a graphical model similar to Mori et al.\cite{Mori2000}, we performed a breadth first search\cite{Cormen2009} to avoid analyses of false airway branches. The removal of false branches is not a trivial task\cite{Mori2000,Irving2014,Kiraly2004}.

We corrected the centreline discretisation error or recentring by smoothing points on the centreline. Smoothing has been an established method in the literature\cite{Irving2014,Xu2015}. A recentring method was proposed by Kiraly et al.\cite{Kiraly2004} which shifts the centreline voxels in relation to a distance transform. The process is iterative compared to a single computation of smoothing.

For our pipeline, we generated the orthonormal plane based on the method of Shirley and Marschner\cite{Shirley2009}. \newtext{We set the pixel size isotopically at 0.3mm to insure that plane image to be within the resolution of the CT image and to allow the ray casting algorithm to find the lumen at sub-voxel precision.} Other methods have been proposed. In Kreyszig\cite{Kreyszig1964}, they generated a binormal and principle normal. However, the method is not robust as the binormal vector can become a zero vector. Grelard et al.\cite{Grelard2015} used Voronoi cells, a method that requires two parameters whereas Shirley and Marschner\cite{Shirley2009} is parameter free. For our work, intensities on the cross-sectional plane were computed via cubic interpolation. Various papers have used linear interpolation\cite{Odry2008b,Kiraly2005,Tschirren2005}. However, it has been shown by Moses et al.\cite{Moses2018} that the method can create high frequency artefacts in the image\cite{Thevenaz2000}.

Various methods have been proposed to measure the area of the airway lumen\cite{Gu2013a,Kiraly2007,Saba2003}. We used the FWHM\textsubscript{ESL} because of two distinct advantages. First, the method is parameter free. Second, the method is robust against slight variations in intensities. The method can therefore be applied to images from different scanners and images acquired using different image reconstruction \newtext{kernels}.

\newtext{\subsection{Limitations}}

In this study, we compared the tapering measurement for healthy and diseased airways using a Wilcoxon Rank Sum Test. The test assumes the data points are independent. However, we used a variety of airways from the same lung. Thus, the tapering profiles of the same patients will have a degree of overlap. Future work is needed to analyse data points that are not dependent on each other.

A key limitation of the tapering measurement is the requirement of having a robust airway segmentation. In this paper, the airway segmentation software was often unable to reach the visible distal point of an airway. Thus, time-consuming manual delineation was needed to extend the missing airways. The distal point is usually located at the periphery of the lungs. Thus, to avoid manual labelling, a segmentation algorithm would need to automatically segment the airways past the sixth airway generation. From the literature, the state of the art software developed by Charbonnier et al.\cite{Charbonnier2017} using deep learning could still only consistently segment airways to the fourth generation. The segmentation of small and peripheral airways is not a trivial task\cite{Moses2018,Bian2018,Yang2018}.

\newtext{In this paper, we analyse the reproducibility of all computerized components of the tapering algorithm. The paper does not address reproducibility of manual labelling of the airways. It is noted in the literature that semi-manual labelling of small airways can take hours\cite{Tschirren2009}. Future work is required to analyse the reproducibility of manual segmentation of the airways. We hypothesise, that the segmented healthy peripheral airways consist of a small number of voxels, therefore any errors in voxel labelling will be considerable smaller then a dilated peripheral airway affected by bronchiectasis.}

\newtext{In this work, we simulated low dose scans through performing Radon transforms on existing CT images, adding Gaussian noise on the sinogram and using backprojection to reconstruct noisy CT images. There are proposed methods to simulate a low dose scans by adding a combination of tailored Gaussian and Poisson noise on the sinogram\cite{Zabic2013}. These methods assume the original high dose sinogram are available for simulation, however it has been acknowledged that sinograms are generally not available in the medical imaging community\cite{WonKim2014,Takenaga2016}. Thus, various groups have proposed low dose simulations using reconstructed CT images. The methods involve adding Gaussian\cite{Takenaga2016,WonKim2014} or a combination of Gaussian and Poisson noise\cite{Naziroglu2017} on the sinogram of the forward projection of the CT image. Whilst there has been limited validation of the appearance of lung nodules against simulated low dose simulation\cite{Li2009}, there has been no validation on the efficacy of these methods on the appearance of airways. We believe that our low dose simulation is sufficient because the measured standard deviation of the trachea mask $T_{n}$ is similar to results taken from low dose scans from Reeves et al.\cite{Reeves2017} and Sui et al.\cite{Sui2016}}

\newtext{Similarly, with voxel size simulation, ideally one would reconstruct the images from the original sinogram, for example in Achenbach et al\cite{Achenbach2009}. However, as the sinograms were unavailable, we simulated the voxel size through interpolation of the original CT images similar to Robins et al\cite{Robins2018}. We believe the simulation is sufficient as it shows the robustness and precision of the centreline, recentring and cross-sectional plane algorithms in the pipeline. Changes in voxel sizes will change the combinatorics or arrangement of the binary image. By showing steps in the pipeline like centreline computation, are repeatable across voxel sizes, we avoid resampling the image to isotropic lengths. Thus, potentially avoiding a computationally expensive\cite{Wiemker2006} pre-processing step.}

\secondnewtext{We showed the tapering measurement is reproducible by measuring the same airway across longitudinal scans with a minimum 5 month interval. The time between scans were on a similar scale from a reproducibility study on airway lumen by Brown et al.\cite{Brown2017}. An ideal experiment to assess reproducibility of the same airway from different scans would be to acquire follow up scans immediately after baseline scans similar to Hammond et al.\cite{Hammond2017}. However, that work was performed on porcine models. Due to considerations of radiation dose, it is difficult to justify the acquisition of additional scans of no clinical benefit\cite{Dendy1999}. For our experiment, each airway was chosen by a subspecialist thoracic radiologist. The airway was inspected to ensure it was in a healthy state, for example, with no mucus present. Thus, we assume that each pair of airways is disease free and healthy.}

\section{Conclusions}

In this paper, we show a statistical difference in tapering between healthy airways and those affected by bronchiectasis as judged by an experienced radiologist. From Figure \ref{Results_tapering_comparision}, the difference between the mean and median of the two populations was 0.011mm$^{-1}$ and 0.006mm$^{-1}$ respectively. In simulated low dose scans, the tapering measurement retained a 95\% confidence interval of $\pm0.005$mm$^{-1}$ up to $\lambda = 3.5$, \newtext{equivalent to a 25mAs low dose scan}. In simulations assessing different voxel sizes, the tapering measurement retained a 95\% confidence between $\pm0.005$mm$^{-1}$ up to $\sigma = 1.5$. The tapering measurement retains the same 95\% confidence, $\pm0.005$mm$^{-1}$ interval against variations in CT reconstruction \replacetext{algorithms}{kernels}, bifurcations and, importantly, over time in evaluating sequential scans in normal airways. \newtext{Importantly, we showed as a proof of principle that the magnitude change in tapering for healthy airways is smaller than those from airways that became bronchiectatic.} \replacetext{Furthermore, we showed that}{From our previous work\cite{Quan2018}, we showed} the measurements are accurate to a sub voxel level. Our findings suggest that our airway tapering measure can be used to assist in the diagnosis of bronchiectasis, to assess the progression of bronchiectasis with time and, potentially, to assess responses to therapy.

We analysed the reproducibility of the components that constitute the tapering measurements. The reproducibility of area measurements was analysed in relation to simulated radiation dose and CT reconstruction \replacetext{algorithms}{kernels}. For simulated dose, we found the 95\% confidence interval retains $\pm 1.5$mm$^2$ in noisy images under $\lambda = 3$,  \newtext{equivalent to a dose just higher than a 25mAs low dose scan}. \secondnewtext{We note on Figure \ref{Dose_noise_area}, there is a bias towards overestimating larger lumen sizes at lower doses. As the centreline length remains constant and bias on the smaller lumen remain stable, the overestimation results in an increase in taper magnitude.} For reconstruction \replacetext{algorithms}{kernels} variation, we found the largest 95\% confidence interval was $\pm 1.9$mm$^2$. The reproducibly of arclengths was tested against voxel sizes variability and showed that arclengths have a 95\% confidence interval of up to $\pm 5.0$mm for scales under $\sigma = 1.5$. The increase in the standard deviation of arclength and area against voxel size and dose respectively correlate with uncertainty in tapering.

This paper provides useful information for clinical practice and clinical trials. \secondnewtext{An accurate prediction of the noise amplitude in a particular CT scan and its distribution is a function of the limited radiation dose of the scan, scanner geometry, reconstructed voxel size, other sources of noise, the reconstruction algorithm and any pre- and post-processing used. Many of these factors are proprietary information of the CT manufacturer and hence not available to users.\cite{Stoel2008,Parr2004} We have undertaken an experiment to assess the dependence of our measurements on a simulated noise field added to the CT scan data and have presented the results. This gives an indication of the dependence on radiation dose assuming all other factors remain the same. We recommend that the accuracy experiment presented in this paper be repeated for the particular reconstruction, scan protocol and scanner type used to make the measurements.}

Bronchiectasis is often described as an orphan disease and has suffered a lack of interest and funding\cite{Hurst2014, Chalmers2014}. We have shown that the reproducibility of automated airway tapering measurements can assist in the diagnosis and management of bronchiectasis.  \secondnewtext{In addition, we show it is feasible to use our tapering measurement in large scale clinical studies of the disease provided careful phantom calibration is taken.} 

\subsection*{Disclosures}
Part of the work have been presented at the 2018 SPIE Medical Imaging conference\cite{Quan2018}.

For potential conflicts of interest: Ryutaro Tanno has been employed by Microsoft, ThinkSono and Butterfly Network (the employment is unrelated to the submitted work), Joseph Jacob has received fees from Boehringer Ingelheim and Roche (unrelated to the submitted work) and David Hawkes is a Founder Shareholder in Ixico plc (unrelated to the submitted work). The other authors have no conflicts of interest to declare.

\acknowledgments 
Kin Quan would like to thank Prof Simon Arridge and Dr Andreas Hauptmann for their helpful conversations on the low dose simulations.

This work is supported by the EPSRC-funded UCL Centre for Doctoral Training in Medical Imaging (EP/L016478/1) and the Department of Health NIHR-funded Biomedical Research Centre at University College London Hospitals.

\secondnewtext{Ryutaro Tanno is supported by Microsoft Research Scholarship.}

Joseph Jacob is a recipient of Wellcome Trust Clinical Research Career Development Fellowship 209553/Z/17/Z 


\bibliography{Med_Phy_cite}   
\bibliographystyle{plain}   


\vspace{2ex}\noindent\textbf{Kin Quan} is a PhD student studying Medical Imaging at University College London. He obtained an MSci in Mathematics from University College London. His research concerns measuring and interpreting the geometry of airways in CT.

\vspace{2ex}\noindent\textbf{Ryutaro Tanno} is a Microsoft Research PhD Scholar. He obtained a BSc and MASt in Pure Mathematics from Imperial College London and University of Cambridge respectively. In addition, he received a MPhil in Computational Neuroscience from University of Cambridge. Ryutaro's interests are designing methods that can cheaply and reliably enhance the quality of medical image data. He has worked as a researcher for Mircosoft, ThinkSono and Butterfly Network.

\vspace{2ex}\noindent\textbf{Rebecca J. Shipley} is a Reader in Biomechanics at University College London. She received a First Class Honours Masters degree (Mathematics), before completing her DPhil (PhD) working on multiscale mathematical models of blood flow and drug delivery in vascularised tissues. Dr Shipley's research involves computational modelling in biology and healthcare with a focus on tissue engineering and cancer. Currently Dr Shipley is Director of the UCL Institute of Healthcare Engineering.

\vspace{2ex}\noindent\textbf{Jeremy S. Brown} \newtext{ is an academic respiratory consultant with a subspecialty interest in lung infection. He trained in respiratory medicine in London alternating with Wellcome funded laboratory research at Imperial College into respiratory pathogens. He has led a laboratory investigating respiratory pathogens at University College London since 2003 and been a clinical consultant at University College Hospitals with a special interest in patients with respiratory infection and bronchiectasis.}

\vspace{2ex}\noindent\textbf{Joseph Jacob} trained as a radiologist at Kings College Hospital with sub-specialist chest radiology training at Imperial College London where he completed an MD(Research). His research interest is computer-based analysis of CT imaging in lung disease for which in 2018, he was awarded a Wellcome Trust Clinical Research Career Development Fellowship. He is based at the Centre for Medical Image Computing and the Respiratory Medicine Department at University College London.\persontext{Bio shorten}

\vspace{2ex}\noindent\textbf{John Hurst} is Professor of Respiratory Medicine at University College London. He has clinical and research interests in bronchiectasis and COPD.  He qualified from the University of Edinburgh Medical School in 1997 and has worked at UCL since 2007.  He has national and international roles with the American and British Thoracic, and European Respiratory Societies and in 2019 will become the Editor in Chief of the European Respiratory Monograph.

\vspace{2ex}\noindent\textbf{David J. Hawkes} is Professor of Computational Imaging Science at UCL. He graduated from Oxford with a BA in Natural Sciences (Physics) in 1974 and obtained an MSc in Radiolobiology in Birmingham in 1975. His research interests are focused on both fundamental research in medical image computing and transfer of advanced computational imaging technologies across the whole spectrum of patient management from screening to diagnosis, therapy planning, image guided interventions and treatment monitoring. \persontext{Bio shorten} \deletetext{He is currently the Director of the Wellcome/EPSRC Centre for Interventional and Surgical Sciences at UCL.}


\listoffigures
\listoftables

\end{spacing}
\end{document}